%% file: main.tex
\documentclass[acmsmall]{acmart}
\makeatletter
\def\@ACM@checkaffil{
    \if@ACM@instpresent\else
    \ClassWarningNoLine{\@classname}{No institution present for an affiliation}%
    \fi
    \if@ACM@citypresent\else
    \ClassWarningNoLine{\@classname}{No city present for an affiliation}%
    \fi
    \if@ACM@countrypresent\else
        \ClassWarningNoLine{\@classname}{No country present for an affiliation}%
    \fi
}
\makeatother

\usepackage{float}
\usepackage{subfigure}
\AtBeginDocument{%
  \providecommand\BibTeX{{%
    \normalfont B\kern-0.5em{\scshape i\kern-0.25em b}\kern-0.8em\TeX}}}
\usepackage{tabularx}
\usepackage{bbding}
\usepackage{graphicx}
\usepackage{geometry}
\usepackage{subfigure}
\usepackage{amsmath}
\usepackage{makecell}
\usepackage{float}
\usepackage{color}
\usepackage{booktabs}
\usepackage{caption}
\usepackage{tabu}
\usepackage{hyperref}
\usepackage{blindtext}
\usepackage{multirow}

\copyrightyear{2025}
\acmYear{2025}
\setcopyright{acmlicensed}\acmConference[CSCW '25]{Proceedings of the}{, 2025}{, }
\acmBooktitle{Proceedings of the}
\acmPrice{15.00}
\acmDOI{xxxxxxxx/xxxxxxxxx}
\acmISBN{xxxxxxxxx/24/04}

\author{Xiaoyu Chang}
\email{changxiaoyu0527@gmail.com}
\orcid{0000-0002-1411-985X}
\affiliation{%
  \institution{Studio for Narrative Spaces, City University of Hong Kong}
  \city{Hong Kong SAR}
\country{Hong Kong}}

\author{Fan Zhang}
\email{zfan1218@gmail.com}
\orcid{0009-0009-9990-8820}
\affiliation{
\institution{Studio for Narrative Spaces, City University of Hong Kong}
\city{Hong Kong SAR}
\country{Hong Kong}}

\author{Kexue Fu}
\email{kexuefu2-c@my.cityu.edu.hk}
\orcid{0000-0002-2929-2663}
\affiliation{%
  \institution{Studio for Narrative Spaces, City University of Hong Kong}
  \city{Hong Kong SAR}
\country{Hong Kong}}

\author{Carla Diana}
\email{cdiana@cranbrook.edu}
\orcid{}
\affiliation{
\institution{4D Design, Cranbrook Academy of Art}
\city{Bloomfield Hills, Michigan}
\country{United States}}

\author{Wendy Ju}
\email{wgj23@cornell.edu}
\orcid{}
\affiliation{
\institution{Information Science, Cornell Tech}
\city{New York, New York}
\country{United States}}

\author{RAY LC}
\authornote{Correspondences should be addressed to LC@raylc.org}
\email{LC@raylc.org}
\orcid{0000-0001-7310-8790}
\affiliation{
\institution{Studio for Narrative Spaces, City University of Hong Kong}
\city{Hong Kong SAR}
\country{Hong Kong}}

\begin{document}
\begin{sloppypar}

\title[A Constructed Response]{A Constructed Response: Designing and Choreographing Robot Arm
Movements in Collaborative Dance Improvisation}

\begin{abstract}
Dancers often prototype movements themselves or with each other during improvisation and choreography. How are these interactions altered when physically manipulable technologies are introduced into the creative process? To understand how dancers design and improvise movements while working with instruments capable of non-humanoid movements, we engaged dancers in workshops to co-create movements with a robot arm in one-human-to-one-robot and three-human-to-one-robot settings. We found that dancers produced more fluid movements in one-to-one scenarios, experiencing a stronger sense of connection and presence with the robot as a co-dancer. In three-to-one scenarios, the dancers divided their attention between the human dancers and the robot, resulting in increased perceived use of space and more stop-and-go movements, perceiving the robot as part of the stage background. This work highlights how technologies can drive creativity in movement artists adapting to new ways of working with physical instruments, contributing design insights supporting artistic collaborations with non-humanoid agents.

\end{abstract}




\maketitle

\section{Introduction}\label{sec:Introduction}
\input{sections/01-Intro.tex}

\section{Related Work}\label{sec:Related Work}
\input{sections/02-Related_Work}

\section{Methods}\label{sec:Methods}
\input{sections/03-Methods.tex}

\section{Results}\label{sec:Results}
\input{sections/04-Results.tex}

\section{Discussion}\label{sec:Discussion}
\input{sections/05-Discussion.tex}

\section{Conclusion}\label{sec:Conclusion}
\input{sections/06-Conclusion.tex}

\begin{acks}
We sincerely thank all dancers who contributed to this research, including Prof. CHAN Anna CY, Prof. SCIALOM Melina, Prof. NGUYEN Ngoc Anh, WANG lu, ZHANG Xuebing, Ming, Li Molin, FAN Jinghang, and Kate. Special thanks to Prof. CHAN Anna CY, Prof. SCIALOM Melina, and Prof. NGUYEN Ngoc Anh for their insights and support for the project.
\end{acks}

\bibliographystyle{ACM-Reference-Format}
\bibliography{main}

\end{sloppypar}
\end{document}

%% file: sections/01-Intro.tex
Dance is an inherently collaborative art form where performers engage in improvisation and choreography, both individually and in groups\cite{yang2024solo}\cite{park2022move}\cite{sandry2017creative}\cite{lubart2021creativity}. These interactions are fundamental to the creative process, allowing dancers to prototype movement designs and refine their performances through continuous feedback and cooperation\cite{otterbein2022dance}\cite{dourish2001}\cite{suchman2007}. Understanding how dancers work with each other and with technology provides insight into the evolving dynamics of creative expression in the performing arts\cite{hsia2021enhancing}\cite{alexander2023using}.

Novel interactions derived from technological advancements offer creative possibilities for dancers\cite{pedersen2020like}\cite{kupca2022creating}. Dancers have long engaged with inanimate objects and props, using them to expand creative expression and explore the relational dynamics between movement and materiality\cite{bennett2020dancing}. These interactions form the basis for integrating modern tools that involve physical interaction, such as robotic arms, into choreographic practice. Robotic systems introduce new modes of collaboration, potentially prompting dancers to adapt their traditional practices to incorporate these technological elements\cite{gomez2021robot}\cite{kim_impact_2020} \cite{huang2018}\cite{lc_active_2023}. Previous CSCW studies have explored dance, design collaboration, and technology. One work explored a technology probe for dancers to decompose movements, demonstrating its use in dance education and practice\cite{riviere2019capturing}. Another work investigates the collaborative dynamics between creators and dancers in contemporary music and dance\cite{hsueh_understanding_2019}. They explored how composers and choreographers interact with dancers through shifting roles and how artifacts mediate these interactions, with role transitions fostering new creative pathways without addressing the integration of robotics or technology in the creative process. Recent work also examines how dancers adapt to virtual performance paradigms\cite{lc2023contradiction}, showing performative strategies such as using technological and time constraints as creative tools and adapting rehearsal workflows to remote interactions. The focus of that work is on remote interactions and virtual settings without addressing collaborative or design processes. The present work addresses the gap in research exploring design processes and creative collaboration between humans and non-humanoid agents in embodied practices like dance. 

\begin{figure}
  \centering
  \includegraphics[width=0.7\linewidth]{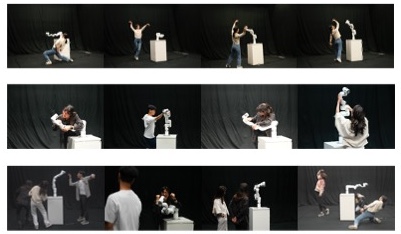}
  \caption{Examples of Dancer-Robot Interactions in Improvisational Dance. One human-one robot improvisation sessions (Top, middle). Multiple humans-one robot sessions from different groups (Bottom).}
  \label{fig:Fig1}
  \Description{Caption}
\end{figure}

To explore how robotic systems affect collaboration dynamics and movement qualities in solo and group settings, we created a workshop with dancers and a non-humanoid robotic arms. We use the intervention to examine how non-humanoid robot presence influences dancers' improvisation, exploring how individual creativity intersects with group collaboration and shared artistic goals. We conducted a series of three interconnected workshops with nine experienced dancers/choreographers to explore how they interact with a non-humanoid robotic arm in different configurations: one dancer with one robot (1-1) and three dancers with one robot (3-1) (Figure 1). Rather than separate engagements, these workshops formed a continuous investigation, allowing dancers to iteratively refine their approaches to movement design and collaboration with the robot. 

The intervention is motivated by the following research questions (RQs):

\textbf{RQ1}: How does a non-humanoid robot affect the way dancers create body movements during improvisation with the robot?

\textbf{RQ2}: How do dancers work in teams to collaboratively improvise and choreograph movements with a non-humanoid robot?

\textbf{RQ3}: How do dancers and choreographers design movements for a non-humanoid robot to facilitate and prototype movement-based performances?

In these workshops, we probed the dancers' experiences and perceptions through semi-structured interviews, observations, and qualitative data analysis. Our findings revealed that dancers produced more fluid movements and felt a stronger connection with the robot in one-to-one scenarios, while group settings led to increased spatial use and more segmented movements, with the robot often perceived as part of the stage background. Our findings highlight how robotic systems can serve as tools for advancing creative choreography and as platforms for end-user development and programming through demonstration, providing actionable frameworks for applications in movement-based learning, interactive education, and collaborative workflows across diverse domains.

This study provided insights into how dancers adapt to and integrate non-humanoid robotic arms into their creative processes, illustrating how technology can drive innovation in movement design and performance. By comparing individual and group interactions with the robotic arm, the study highlights differences in movement quality and dancer engagement, offering implication for design of future robotic systems that support artistic expression. Additionally, we emphasize the importance of interdisciplinary collaboration, showing how knowledge transfer between dance and robotics can lead to creative possibilities for augmenting human creativity\cite{hartmann2006}\cite{schmidt2002}.

%% file: sections/02-Related_Work.tex
\subsection{Collaboration in Dance}
Previous work have explored the dynamics of collaboration in dance, including choreography, co-improvement in the dance-making process, and collective improvisation.

Various studies have noticed the collaborative nature of choreography as dancers and choreographers closely work together to create dance pieces. Rowell \cite{grau_united_2000} noted an emerging new status for dance as a collaborative art through an analysis of contemporary dance in Europe and the UK. Klien \cite{silvia_broecker_choreography_2007} also stated that dancers in the current political context have to balance their individual freedom and personal experiences within a larger group instead of simply being hired to perform. In the similar context of UK’s contemporary dance, Butterworth \cite{butterworth__teaching_2004} identified five distinct collaborative processes between choreographers and dancers, and introduced the \textit{Didactic-Democratic Spectrum} framework, which outlines varying levels of collaboration between dance artists. As Gibbons \cite{gibbons_co-authorship_2015} later described, choreographers and dancers collaborate to generate materials and inspire each other, especially in the "editing process". In this dance making process, the choreographer can be viewed as a "curator" who organizes, compiles, and arranges choreographic objects, movement material and structures. Carroll et al. \cite{carroll_bodies_2012} demonstrated how the choreographer collaborated with dancers using \textit{Choreographer’s Notebook} during the dance production process. The hierarchical nature of this collaboration was also revealed, as only choreographers provided feedback on dancers' videos, while dancers did not correct each other. Recently, Ciolfi Felice et al. \cite{felice_studying_2021} presented a deployment of a creativity-support tool Knotation in a long-term study to explore the collaborative process and shifting roles between choreographers and dancers in dance making.

Dance often consists of following an internal (the dancer's own feelings and emotions) or external rhythm (movements of the partner or a group), as well as coordination with the space \cite{passos_interpersonal_2016}. In many dance styles, dancers must practice interpersonal coordination, managing time and space, and varying their movements to synchronize or contrast with their partners. For example, when performing contact improvisation, which involves two dancers maintaining physical contact and responding to each other’s movements \cite{buckwalter_composing_2010}, the dancer acquires kinesthetic interconnectivity to sense the partner's body and the ground \cite{kimmel_sources_2018}. They respond to the very moment naturally through specific skills like "listening", and creating with the moving body to the largest extent. In this process, dancers learn to track their partner's attention and even decide the direction of movements based on tactility, weight distribution, gaze, and temperature sensed by their partners. Historically, dancers have also worked with inanimate objects or props\cite{dong2024dances}\cite{bennett2020dancing}, to enhance creative expression and explore spatial dynamics. These interactions serve as precursors to understanding the dancer-object relationship, offering valuable insights into how technology like non-humanoid robots could further enrich collaborative and creative practices in dance.

Prior studies offer related but distinct insights into the intersections of dance, collaboration, and technology. Hsueh et al.\cite{hsueh_understanding_2019} examined collaborative relationships between creators and dancers in contemporary music and dance, emphasizing role transitions but not addressing robotic or technological integration. Rivière et al.\cite{riviere2019capturing}introduced a technology probe to assist dancers in decomposing movements, highlighting the role of technology in dance education and practice without focusing on collaborative or design dynamics. Our study bridges these gaps by examining how non-humanoid robotic systems influence creative processes in dance, emphasizing embodied interaction, kinaesthetic creativity, and collaborative choreography. By exploring in-person, real-time dance collaboration dynamics with robotic technologies, our work expands the understanding of how humans co-create with non-humanoid robots in performance-driven contexts.

\subsection{Technology-mediated Movement Interactions}

Integrating technology into dance has been proven to influence movement creation, teaching, and performance. For example, Motion capture (MoCap) systems, provide dancers with real-time feedback, allowing them to visualize and refine their movements dynamically during rehearsals\cite{carlson2021mocap}\cite{strutt2022motion}, and mocap avatars may affect the quality of dance movements \cite{zhang_becoming_2025}. Virtual reality (VR) creates immersive environments, enabling novel forms of dance expression and audience interaction\cite{alaoui2021rco}\cite{bouchard2022immersive}\cite{cao_dreamvr_2023}\cite{fu_i_2023}, while haptics provide diverse interactions for performers\cite{shen_its_2025}. These technologies enhance dancers' engagement with the art form, fostering deeper connections with dancers' movements and audience.

Robotic systems in dance move beyond tools to become active collaborators in performances\cite{castro2024body}\cite{lc_presentation_2022}. Research on co-creative processes between humans and robots in contemporary dance shows how robots bring unique capabilities and challenges to the performance space\cite{zhou2021dance}\cite{berthaut2019co}\cite{lc_active_2023}. Our study extends existing research on end-user development and programming through demonstration, particularly in the context of human-robot collaboration, where users interactively design and adapt robotic behaviors. Similar to the methods employed in prior works\cite{dong2024dances}\cite{alaoui2021rco}, our workshops involved choreographers programming robotic arm movements through demonstration, enabling a collaborative environment that fosters creativity, learning, and adaptation. This approach aligns with interdisciplinary applications of robotic systems in educational settings and creative industries, where users actively engage in shaping the technology to support their artistic and functional objectives. Our research highlights how dancers and choreographers adapt their practices, using the robot's distinct physicality and movement patterns to inspire new choreographic ideas\cite{castro2024body}.

Robotic arms represent a fascinating blend of precision and versatility in dance\cite{rogel2022robogroove}\cite{rogel2022music}. Dancers adapt their choreography to the mechanical properties of robotic arms, exploring the creative potential that arises from this interaction\cite{kim_impact_2020}. Unlike human dancers, robotic arms provide consistent, repeatable movements, enabling intricate synchronization and complex patterns that enhance the visual and emotional impact of performances. This consistency pushes dancers to innovate within the robot's capabilities, finding new ways to express their artistry.

Research on the integration of robotics and interactive technologies in dance has shown how these robots can enhance collaborative creativity\cite{zhang2023ai}. Studies have also explored how haptic feedback and interactive lighting synchronized with robotic movements create multi-sensory dance experiences\cite{lee2022synchronizing}. These advancements illustrate the evolving dynamics of creative expression in the performing arts, emphasizing the importance of understanding how dancers work with new instruments to enrich the art\cite{lc_power_2022}. Our study builds on prior work in end-user development and programming through demonstration, emphasizing human-robot collaboration in creative settings. Similar to methods explored in prior studies\cite{dong2024dances}\cite{alaoui2021rco}, choreographers in our workshops programmed robotic arm movements through direct demonstration. This enabled a collaborative process that supports creativity, iterative learning, and adaptation, aligning with interdisciplinary applications in educational and creative industries. Furthermore, while existing research predominantly focuses on tools like MoCap, VR, and humanoid robots, the role of non-humanoid robotic systems remains underexplored. By integrating robotic arms with unique movement capabilities, our work examines how such technologies challenge traditional collaboration paradigms and inspire dancers to innovate their creative workflows.

\subsection{Perception and Interaction with Non-humanoid Robots}

Non-humanoid robots, designed without mimicking human form, play functional and task-specific roles in various domains\cite{berka2018humanoid}\cite{rogel2022music}\cite{chang2024sorry}, including the performing arts. Unlike humanoid robots, which aim to replicate human behaviours and interactions\cite{duffy2003anthropomorphism}\cite{mori2012uncanny}, non-humanoid robots like robotic arms offer unique opportunities and challenges in performance settings. 

Non-humanoid robots, designed for specific functions, influence how humans perceive and interact with them. Their distinct forms and movements shape these interactions, with smoother, and more predictable motions enhancing comfort and acceptance\cite{takayama2011expressing}. This predictability is crucial for dancers, as it facilitates better collaboration and understanding, allowing dancers to anticipate and synchronize with robotic movements\cite{dragan2012generating}. The interaction with non-humanoid robots, therefore, is faster more intuitive and engaging collaborations.

In performance arts, non-humanoid robots like robotic arms offer novel forms of artistic expression\cite{berthaut2019co}, enable complex choreographies that challenge traditional dance boundaries\cite{abe2022beyond}. These robots provide consistent, repeatable movements, pushing dancers to innovate within the robot's capabilities. Existing research has primarily emphasized the practical and aesthetic contributions of humanoid robots to performance arts, often overlooking the creative and collaborative dynamics of non-humanoid systems. This work bridges that gap by focusing on non-humanoid robotic arms and their potential to act as co-creators in dance. Our work highlights how the lack of anthropomorphism in robotic systems encourages dancers to explore alternative movement vocabularies, fostering unique choreographic approaches and expanding the scope of artistic expression.

\subsection{Study on Choreographers and Dancers}

Movement-based collaboration with technology has been extensively examined, focusing on how these technologies can enhance artistic expression and performance capabilities. This field investigates the collaborative dynamics between human dancers and robotic systems, offering insights into the augmentation of traditional dance through technological means.

Choreographers have increasingly viewed robotic technologies as tools that extend the boundaries of traditional dance. These technologies facilitate the exploration of novel choreographic possibilities and enable the creation of intricate and precise movement patterns. Previous research elucidates how digital performance technologies have revolutionized contemporary dance, opening new creative avenues\cite{dixon2007digital}. And the collaborative potential between human dancers and robots, highlighting the co-creation of dance performances\cite{berthaut2019co}. For example, dancers adapt both physically and cognitively when working with robots. This adaptation involves understanding robotic movements and programming, thereby developing a hybrid skill set that integrates dance and technology. When dancers interact with robots\cite{calvo2005action}, they use parts of their brain that help them observe actions and use their motor skills.

Robotic technologies enhance artistic expression and emotional engagement in dance performances. Robots can be programmed to execute movements that evoke specific emotions, thus adding depth to the performance. The aesthetic interplay between humans and machines has been discussed, creating unique artistic experiences that challenge conventional dance paradigms \cite{kozel2010cloth}. This technological-artistic fusion not only expands the scope of dance but also enriches the emotional and aesthetic experiences of both dancers and audiences.

The integration of robots into dance and choreography poses challenges, including the necessity for technical proficiency and potential creative constraints due to technological limitations. The complexities of synthesizing self-organized dance with robots are mentioned, emphasizing the importance of interdisciplinary collaboration between artists and technologists \cite{bremner2018robot}. However, these challenges also present opportunities for innovation, as the precision and reliability of robots can inspire new forms of movement and interaction, thereby pushing the boundaries of traditional dance \cite{kim_impact_2020}.

Enhancing the collaborative dynamics between humans and robots in dance includes developing interfaces for programming robotic movements and exploring new interaction forms that leverage both human creativity and the robotic. Long-term studies on the impact of robotic integration on artistic practices will provide deeper insights into the evolving relationship between technology and art. Understanding how dancers and choreographers adapt to and innovate with robotic technologies will be crucial for advancing this interdisciplinary field.

However, little research has addressed how choreographers and dancers adapt their creative practices with non-humanoid robotic systems, especially in group settings\cite{hsueh_understanding_2019}\cite{riviere2019capturing}. Our study explores the collaborative dynamics and movement designs between one-on-one and group interactions with robotic arms. By focusing on how non-humanoid robots influence creative workflows, our work provides qualitative insights that contribute to interdisciplinary collaboration in performing arts and inform the design of robotic systems that support artistic innovation.

%% file: sections/03-Methods.tex
\begin{figure}
  \centering
\includegraphics[width=0.6\linewidth]{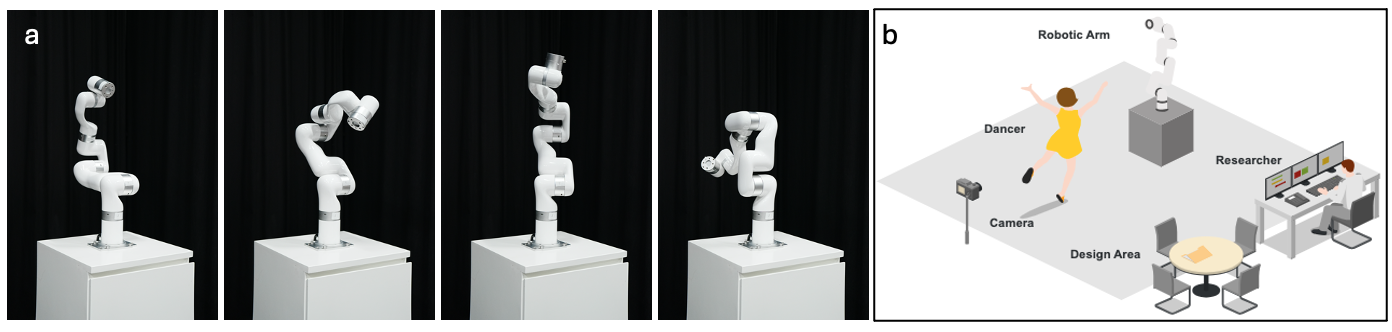}
  \caption{Experimental Setup Overview: a) Non-humanoid robotic arm used in the workshops. b) Schematic of the experimental setup.}
  \label{fig:Fig2}
  \Description{Caption}
\end{figure}

\begin{figure}[htbp]
  \centering
  \includegraphics[width=0.8\linewidth]{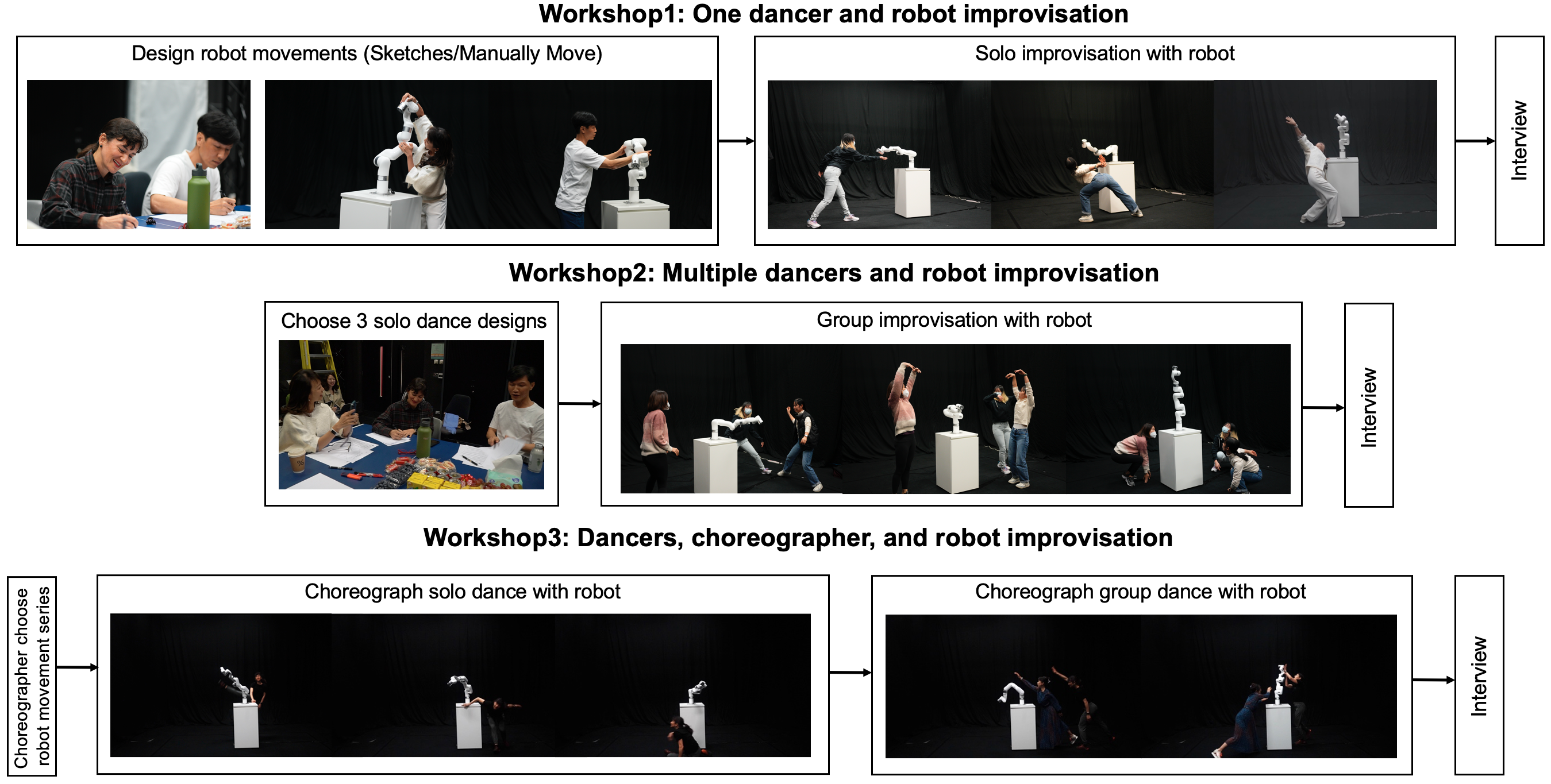}
  \caption{Co-Designing Robot Arm Movements with Dancers.
Top: dancers individually design and perform solo with robot movements. Middle: Group collaborates to synthesize individual designs and perform with the robot. Bottom: One choreographer creates a set of robot arm actions for a single dancer or two other dancers, who then perform with the robot.}
  \label{fig:Fig3}
  \Description{Caption}
\end{figure}

\begin{table}[]
    \small
    \caption{Dancers in this Work}
    \label{tab:dancers_this_work}
    \centering
\begin{tabular}{llllll}
\hline
Group & Dancer & Gender & Age   & Dance Category     & Dance Years \\
1     & P1     & Female & 41-50 & Swing              & 20          \\
      & P2     & Female & 31-40 & Hip-hop            & 9           \\
      & P3     & Female & 21-30 & Street Dance       & 7           \\
2     & P4     & Female & 51-60 & Improvisation      & 40          \\
      & P5     & Male   & 41-50 & Ballet             & 35          \\
      & P6     & Female & 41-50 & Contemporary Dance & 30          \\
3     & P7     & Female & 21-30 & Modern Dance       & 16          \\
      & P8     & Female & 21-30 & Modern Dance       & 20          \\
      & P9     & Female & 21-30 & Contemporary Dance & 5   \\
\hline       
\end{tabular}
\end{table}

\subsection{Research Setting}

Each session was conducted in a controlled laboratory environment. An xArm 6 robotic arm (UFactory, Shenzhen) was centrally placed on the floor. The robotic arm stood on a white platform with dimensions of 150 cm in height, 50 cm in length, and 50 cm in width, allowing dancers to move around it with a proximity ranging between 0.5 m and 1.5 m. The experimental setup consisted of a non-humanoid robotic arm designed for improvisational dance studies and an arrangement tailored for three workshops. As shown in Figure 2, panel a illustrates the robotic arm, and panel b provides an overview of the setup layout.

\subsection{Participants}

We organized three separate participation sessions with three groups. We recruited nine professional dancers (eight female, one male; ages 21–60) from a local academic dance university using purposive sampling, selecting participants based on their professional experience and expertise in dance. The nine dancers were randomly assigned to three groups, each participating in three half-day workshops. As detailed in Table 1, these dancers had between 5 and 40 years of experience in various dance styles, ranging from swing to contemporary dance. The dancers did not have prior experiences dancing with the robot or robotic arm, with 2 had prior experience with drones. All the dancers had prior experience dancing with inanimate objects like chairs. Three researchers conducted data collection, facilitated workshops, and recorded observations. One researcher managed participant coordination and guided interviews. Another operated the robotic system, ensuring consistency in programmed movements. The third researcher focused on documentation, including video recording and note-taking. All researchers followed a structured protocol to minimize bias and maintain consistency in data collection. The workflow for these sessions is depicted in Fig. \ref{fig:Fig3}. 

For Workshop 1, each dancer was asked to manually move the robotic arm first to become familiar with the robotic arm dynamics modes, so that they would not conflict with the movements of the robotic arm in future steps, ensuring their safety. Then each dancer did the movement design for the robotic arm, and the researcher played the movement design recording for the dancer to do the solo improvisation. 

For Workshop 2, each group had a discussion first to choose the movement design from Workshop 1 and then did group improvisation with the robotic arm conducting the chosen movement design. 

For Workshop 3, each group collaboratively decided to adopt the roles of one choreographer and two dancers for their work. The choreographer chose three movement designs in Workshop 1, which were then combined into new movement series by the researcher. The choreographer created a solo dance sequence for one of the two dancers to perform with the robotic arm. And then two dancers performed together with the robotic arm following the choreographed duet performance sequence. This organization ensured that each group had a balance of creative direction and performance capabilities.

These workshops were designed to let the participants think from the sides of designing movements, choreographing movements and improvisation based on their interactions with the robotic arm.

Before the experiment, all participants signed a consent form, which explicitly granted permission for their faces to be shown in photographs, the use of their data, and the publication of these images as part of the study. All procedures followed the ethical guidelines established by our Institutional Review Board (IRB).

\subsection{Technology Implementation}

In our series of workshops, the UFactory xArm 6 robotic arm was central to exploring the intersection of non-humanoid robotics and dance. The system comprised the robotic arm, Ufactory's control software, a laptop for recording movement design and control, and an external speaker for music play. Each session utilized the same non-lyric music, "Mo Better Blues," to ensure a consistent auditory environment.
\begin{figure*}[htbp]
  \centering
  \includegraphics[width=0.8\linewidth]{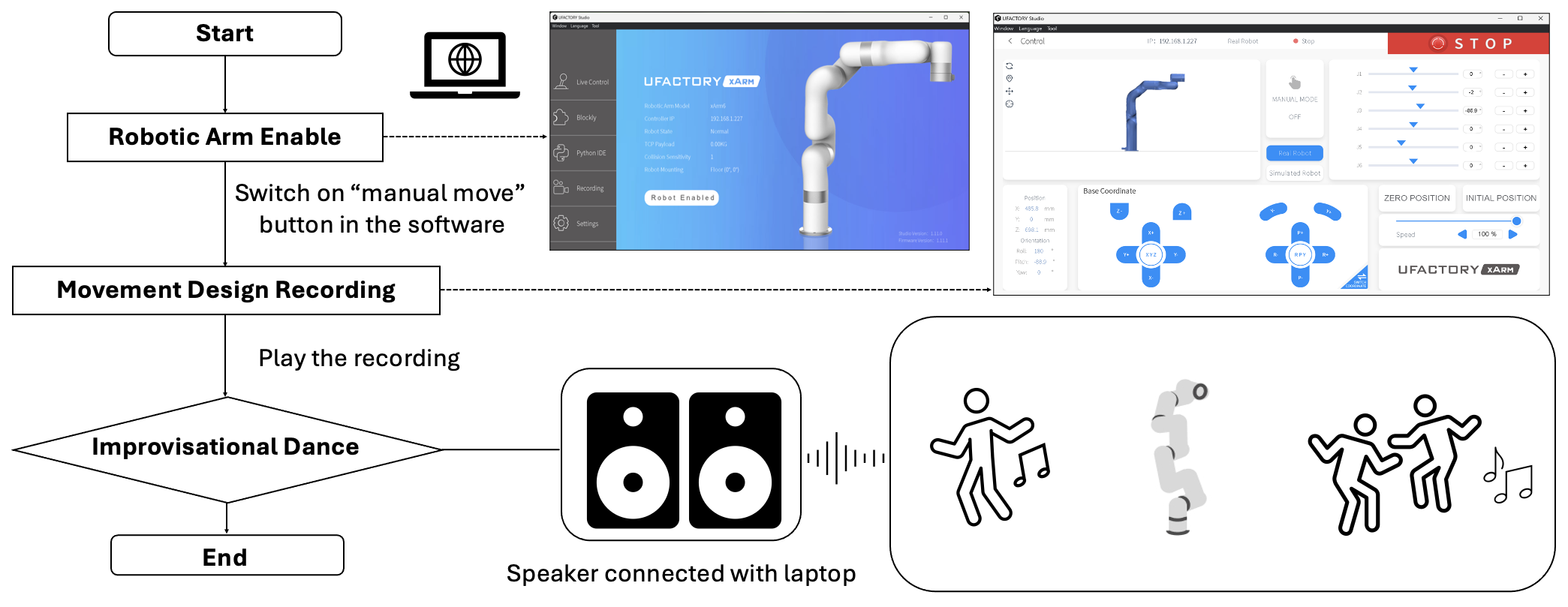}
  \caption{System Overview for Robot Control and Movement Recording:  The robotic arm is controlled through Ufactory software connected to a laptop. The system enables movement design and recording by selecting "manual move" and then clicking "recording." The recorded movement is played back for dancers' improvisational performance, with music played through a connected speaker.}
  \label{fig:Fig6}
  \Description{Caption}
\end{figure*}

The robotic arm was initialized using Ufactory's control software. This involved checking all joints and movement axes for proper alignment and functionality, a crucial step to guarantee the precision required for the workshops. Then, dancers engaged with the robotic arm through the software's "manual move" button. This mode allowed them to manipulate the arm physically, creating their movement sequences. Once dancers started their movement designs, the software's "recording" function was activated. This feature captured each movement, ensuring fidelity in playback. The recording process involved capturing the exact trajectory and timing of the movements, which were later used for both solo and group improvisational sessions. During the improvisational dance, multiple recorded sequences were played in a predefined order executed by the robotic arm in synchronization with the music (Fig. 4). 

\subsection{Data Acquisition}
To systematically collect and analyze data, we implemented several approaches involving video recording, photography, sketch scanning, note-taking, and semi-structured interviews. This data acquisition strategy was designed to capture a wide range of qualitative data to address our RQs thoroughly. 

A camera was placed on a tripod in front of the setup to capture the interactions between dancers and the robotic arm (Fig.2b). Photographs were taken to document significant moments and details of the experimental setup and participants' interactions. Participants' sketches, which were part of the movement design process, were scanned to preserve the visual data for later analysis. Researchers also maintained detailed notes during each session to record observations and spontaneous insights, providing a rich qualitative layer to complement the visual data.

In addition, semi-structured interviews were conducted after each workshop to align closely with the research questions and gain insights into participants' experiences. The interview questions were tailored to the unique focus of each workshop:
\textbf{Workshop 1:} Questions explored participants' perceptions of their relationship with the robotic arm (e.g., leader/follower roles), how the robotic arm influenced their movement decisions, and their creative interaction process.
\textbf{Workshop 2:} Questions examined participants' experiences in group improvisation (3-1) compared to solo settings (1-1). Themes included:
Perceptions of collaboration dynamics: "What are the major differences between 1-1 and 3-1 settings?"
Spatial exploration: "When do you explore space more, 1-1 or 3-1? Why?"
Attention distribution: "During 3-1, which do you pay more attention to, the robotic arm or your partner dancers? Why?"
Connection with the robotic arm: "Which case makes you feel more connected to the robotic arm?"
Creative inspiration: "Did the robotic arm inspire you during the dance?"
These questions provided insights into how group settings influenced spatial dynamics, attentional focus, and dancers' engagement with the robotic arm.
\textbf{Workshop 3:} Questions centered on movement generation and transformation, focusing on the rationale for selecting specific motifs, adapting choreography for the robotic arm, and comparing human and robotic movement designs.

\subsection{Workshop 1: Robotic Rhythms: One Dancer Exploring Dance Movements with One Robotic Arm}
Workshop 1 examined how individual dancers interact with a non-humanoid robotic arm during movement design and improvisation(see Fig.3).

The session included four phases:

1. Demonstration:
Participants observed the robotic arm performing a pre-defined dance sequence called "Born" movements (Fig. 4)\cite{lc2023contradiction}, providing a reference for the robot's movement style.

2. Movement Design:
Each dancer created three sequences using sketch ideas and physically guiding the robotic arm's joints. During this process, dancers acted as both robot programmers and interactors—designing movements and later dancing alongside the robot executing those motions. A researcher served as the robot executor, recording and replaying the movements for interaction (Fig. 5). 

Before programming, researchers emphasized safety precautions and explained the rigidity of the robotic arm. The hands-on programming phase familiarized dancers with the robot’s physicality and helped them intuitively maintain safe distances during interaction.

3. Improvisation:
Each dancer then performed an improvised solo alongside the pre-programmed robotic arm to explore dynamic co-creation and embodied responses.

4. Post-Session Interview:
Dancers participated in 20-40 minute semi-structured interviews following the session to share their experiences and insights.

\subsection{Workshop 2: Mechanical Motifs: Multiple Dancers Creating Dance Concepts with One Robotic Arm}
Workshop 2 was structured to facilitate collaborative creation and interaction between multiple dancers and the robotic arm, where the same nine professional dancers from Workshop 1 explored group interaction with the robotic arm (see Fig.3), following a similar engagement approach as outlined in the referenced study\cite{hsueh_understanding_2019}. 

The session included: 
1. Group Formation and Discussion:
Dancers were divided into three groups. Each group selected three movement designs from the nine movements generated in Workshop 1 and collaboratively planned a cohesive movement sequence using them.

2. Group Improvisation:
Each group performed an improvisational dance with the robotic arm. This workshop stage emphasized collective creativity and coordination, exploring how multiple dancers interact with and respond to the robotic arm's movements and how dancers interact with their human partner.
To clarify the roles during this process, one dancer served as the robot programmer, designing the movement sequence for the robotic arm. Another dancer acted as the robot interactor, performing the dance while the robotic arm executed the pre-designed movements. A researcher played the role of the robot executor, recording and playing back the designed movements for the dancers to interact with.

3. Post-Session Interviews:
Participants reflected on the group dynamic, choice of movements, and the role of the robotic arm, using the same semi-structured interview format as workshop 1. 

\subsection{Workshop 3: Automated Artistry: Generating Dance Movements with Robotic Arms}
Workshop 3 examined choreographed dance movements involving solo and duet interactions between dancers and a robotic arm (see Fig.3).

1. Movement Series Creation:
Each choreographer selected and combined three movement designs from Workshop 1 to form a new dance sequence.

2. Solo Performance Choreography:
The choreographer first created a dance sequence for one dancer to perform with the robotic arm.

3. Duet Performance Choreography:
Following the solo performance, the choreographer developed a duet dance sequence involving two dancers and the robotic arm.
To clarify the roles during this process, one dancer served as the robot programmer, designing the movement sequence for the robotic arm. Another dancer was the robot interactor, performing the movements while the robotic arm executed the pre-designed sequences. A researcher served as the robot executor to record and play back the choreographed movements for the dancers to perform with.

4. Post-Session Interviews:
Choreographers and dancers participated in a semi-structured group interview to discuss their perceptions and experiences of dancing with the robotic arm in a choreographed setting compared to the improvisations explored in Workshops 1 and 2. The interview focused on understanding how the role of the robotic arm was perceived in the context of structured choreography and how this influenced their creative process and performance dynamics.

\subsection{Data Analysis}
Interview transcripts were independently coded by researchers using thematic analysis guidelines\cite{braun2006using}. Each researcher identified preliminary codes, which were then collaboratively refined into broader themes through iterative discussion. This process ensured analytical consistency and grounded the results in participants’ perspectives. The organization and interpretation of these themes provided insights into the dancers' experiences and perceptions, which are presented in detail in Section 4.

Video recordings were coded to capture key moments of interaction, such as the use of space, shifts in attention, and movement quality. These observations provided visual evidence that contextualized the interview data.  

Sketches created during movement design were analyzed to understand participants’ conceptualizations of robotic motion. These sketches were compared to interview narratives and video observations, highlighting connections between dancers’ intentions and their embodied practices.

Photographs, used primarily for documentation, supplemented the video and sketch analysis by clarifying spatial configurations and notable interactions referenced during interviews.

We employed an affinity diagram approach to synthesize insights from all data\cite{iqbal2022systematic}\cite{awasthi2012hybrid}. Themes around attention distribution in group sessions were validated through matching video cues and spatial arrangements depicted in sketches.

\begin{figure}[htbp]
  \centering
  \includegraphics[width=0.75\linewidth]{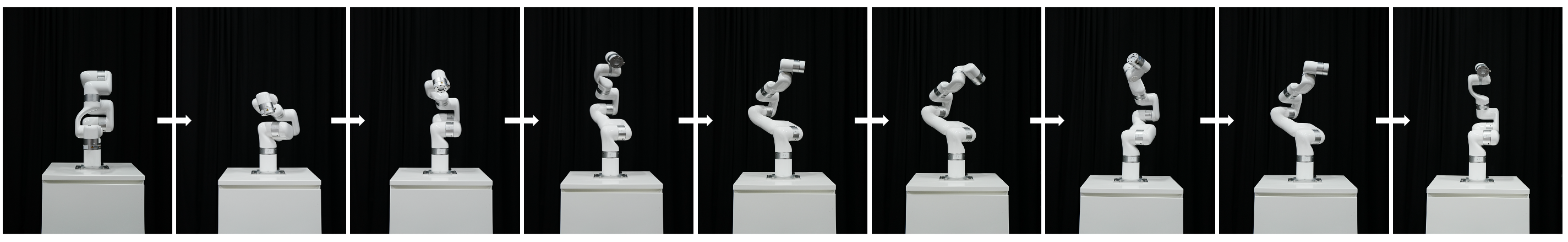}
  \caption{Movement Design demonstration vignettes: "Born" movements series in \cite{lc2023contradiction}. }
  \label{fig:Fig4}
  \Description{Caption}
\end{figure}

\begin{figure}[htbp]
  \centering
  \includegraphics[width=0.45\linewidth]{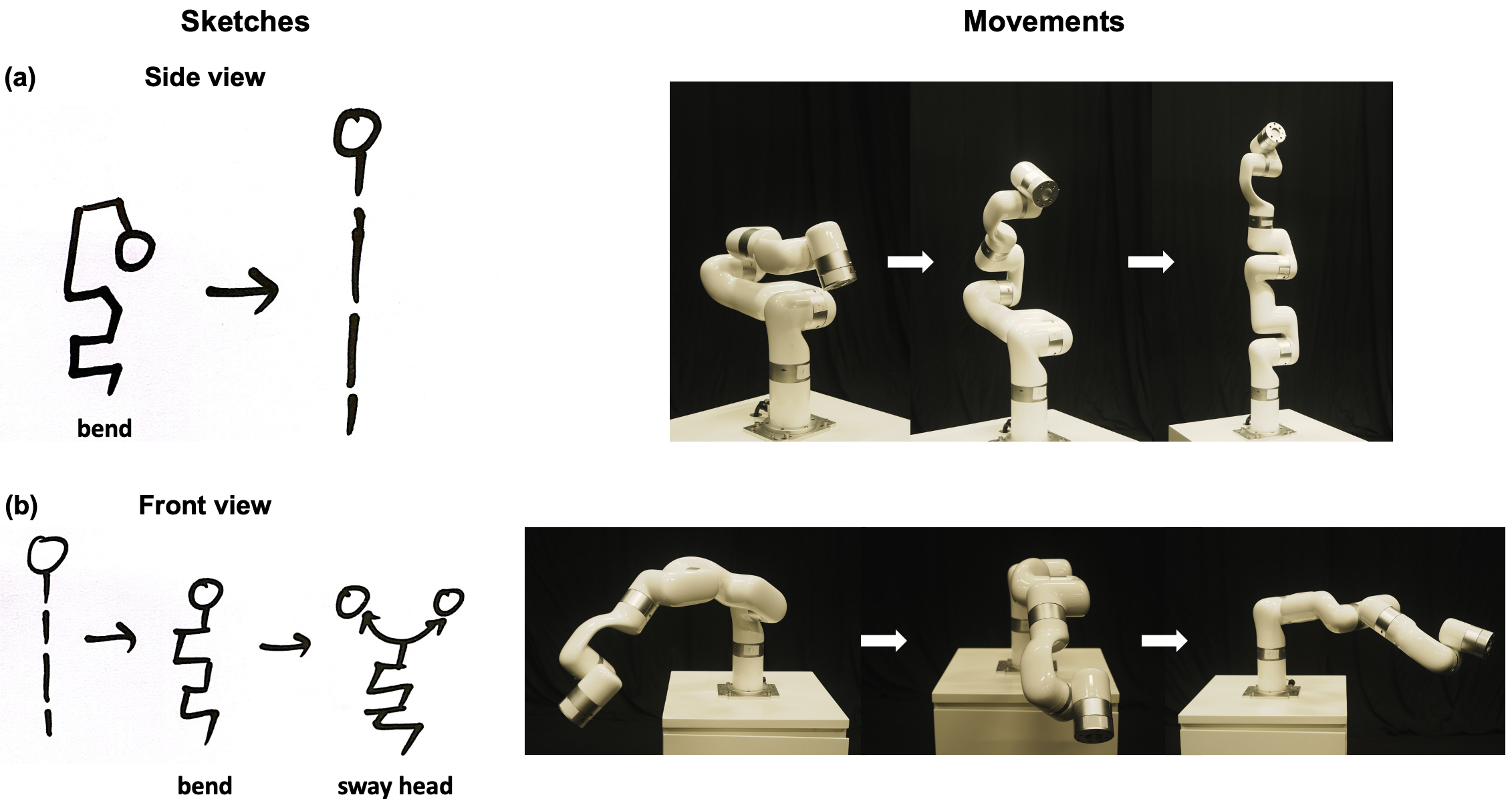}
  \caption{Sketches and Programmed Movement Designs: 
Left: Movement designs created by professional dancers.
Right: Recorded robot movements based on dancers' designs.}
  \label{fig:Fig5}
  \Description{Caption}
\end{figure}

%% file: sections/04-Results.tex
We explored the impact of varying numbers of dancer partners, delved into their perspectives on the role the robotic arm plays, how the dancers distributed their attention, spatial interactions and explorations, and highlighted distinctive attributes that set the robotic arm apart from human partners. We also explored how the robotic arm influenced the choreographer's movement generation and the dancer's perceptions in choreographed vs. improvisational dance. Qualitative insights are detailed in this section. 

\subsection{Spatial Kinematics and Interactive Dynamics with Robotic Arm}


In solo sessions, dancers led the robotic arm in continuous movements; in group settings, their movements became more discontinuous. Dancers felt free and predictable in solo sessions but experienced spatial constraints and spontaneity in group settings. Across all workshops, spatial awareness and exploration, navigation hazards, visual balance, and stage utilization were key considerations. 

\subsubsection{Distinct Movement Improvisations in Individual vs. Group Performance.}

The dynamics of leading and following between dancers and the robotic arm and among human dancers revealed distinct patterns during the improvisations. 
In individual sessions, dancers experienced a clear sense of control and leadership over the robotic arm. P5 described the robot as "a toy, an object" with "higher intimacy" in the one-on-one setting. P3 supported this by stating, "Lead, because I need to interact with the robotic arm, and I need to do the performance with it and consider the visual effects." This sense of manipulation allowed dancers to feel as if they were leading the interaction, designing and controlling the arm's movements. 
In contrast, group improvisation involved more complex dynamics. Dancers navigated their interactions with both the robotic arm and their human partners. P2 explained, “I have more attention on the humans…P1 gives us hints.” The robotic arm's presence became less central, and dancers often followed human cues over robotic ones.

\begin{figure}[htbp]
  \centering
  \includegraphics[width=0.8\linewidth]{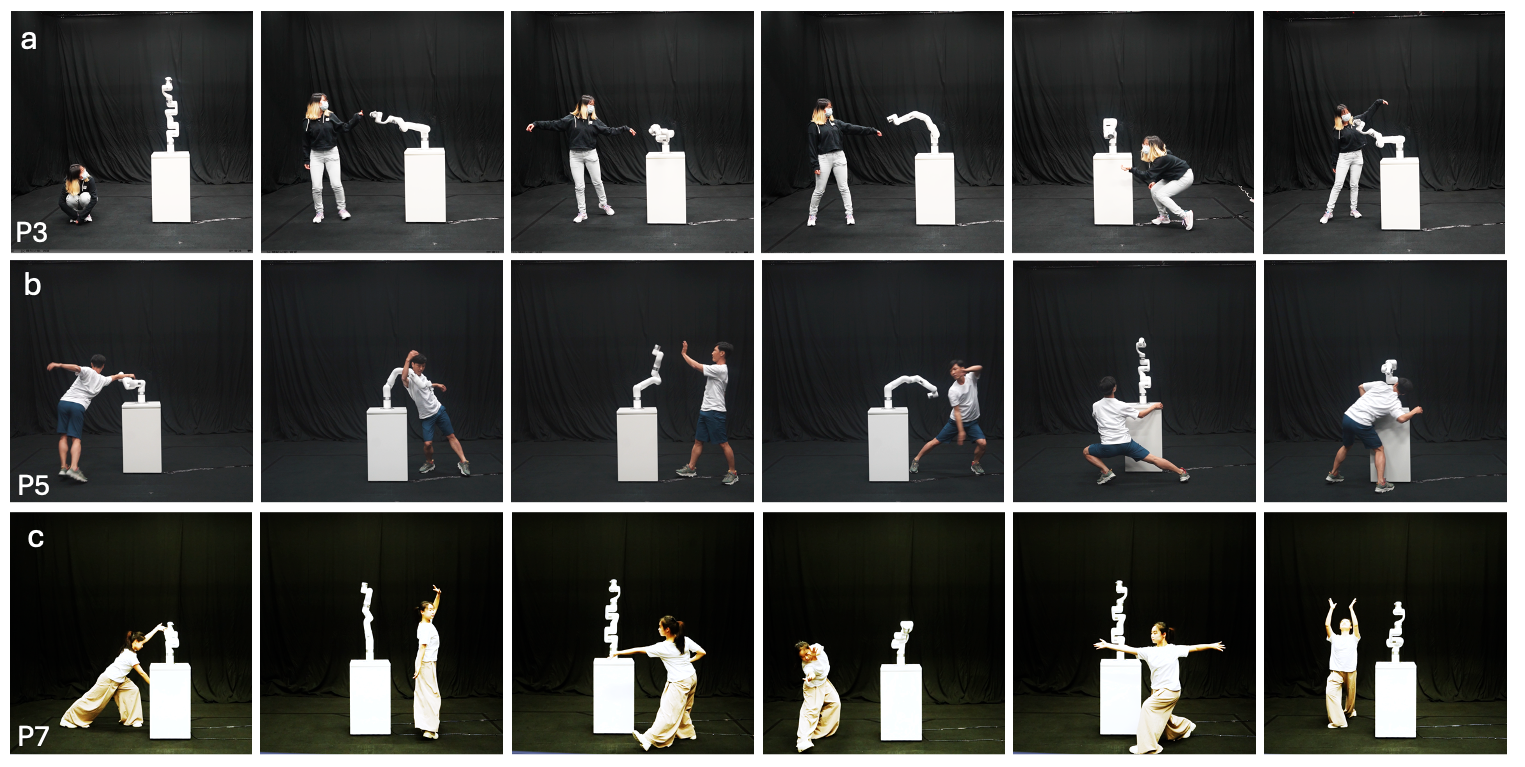}
  \caption{Examples of dancers' solo improvisational dance. a. P3, b. P5, c. P7 conducted continuous and intimate individual improvisational dance with the robotic arm.}
  \label{fig:Fig7}
  \Description{Caption}
\end{figure}

Movements in solo sessions were continuous and fluid (Fig. 7), reflecting a deeper engagement with the robotic arm. P5 noted the coherent connection of the session: "When I did 1-1, I had to do a lot of intimacy and stay with the robot.” While in group improvisation, the dynamic shifted to more discontinuous movements (Fig. 8). P5 stated, “3-1 is interacting with space and each other, but 1-1 somehow, I want to try to strip out, but I still need to draw back to the robot because it just fits into space in the centre. But with the human, I was able to step out to look at these two ladies dancing, then I joined, and then I took out and walked around.” The necessity to interact with multiple dancers introduced interruptions and variations in the flow. P5 highlighted this by saying, "With more people in the space, I do that option around. I stop and restart." “When I feel something happening, probably, I see more people behind me and then they're moving forward, I stop, and then I leave the space." This discontinuity reflected the need to constantly adapt to the changing inputs and actions of the group, contrasting the more predictable and steady interaction with the robotic arm in individual sessions.

In the individual improvisation, dancers reported a higher level of connection with the robotic arm. P5 described the one-on-one sessions as "more intimate," where the focus was on a direct and personal connection with the robot by noting, "When I did 1-1, our attention is directly on the robot. The robot’s movements are designed by us, so we clearly know the structure, the timing, and the movements, so we are very intimate, and the relationship is more intimate." The dancers' constant and exclusive interaction with the robotic arm made it easier for them to develop a deeper connection with it.
However, in group improvisation, the sense of intimate connection was diluted. The presence of multiple human partners shifted the focus away from the robotic arm, reducing the direct and personal connection. P5 observed, "3-1 is more variation and the level thing. When I did 1-1, I had to do a lot of intimacy stay with the robot." The group's dynamics introduced a broader range of interactions, making the relationship with the robotic arm with fewer connections.

\begin{figure}[htbp]
  \centering
  \includegraphics[width=0.8\linewidth]{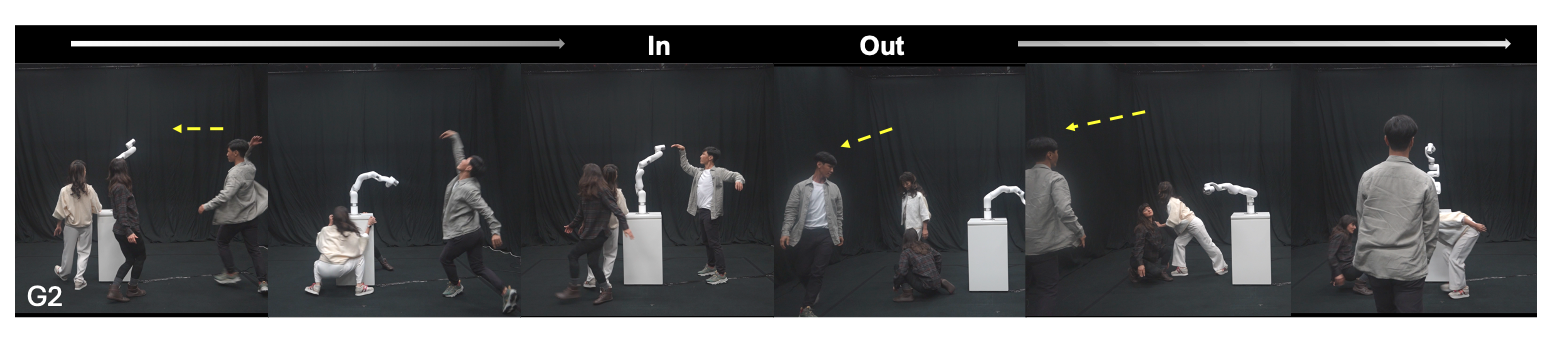}
  \caption{Discontinuous and non-intimacy in group improvisational dance. P5 is doing "In and Out" when doing group improvisational dance with the robotic arm and human dancers.}
  \label{fig:Fig8}
  \Description{Caption}
\end{figure}

This finding highlights the distinct experiences in individual versus group improvisation. In individual sessions, dancers led the robotic arm, engaging in continuous movements and intimate, controlled connections. In group settings, the need to coordinate with human partners shifted the dynamics toward following, resulting in more fragmented movement and a weaker connection with the robot. These differences in control, continuity, and engagement offer valuable insights into human-robot collaboration.

\subsubsection{Transition From Constructed Design Responses to Improvisational Movements.}
In Workshop 1, dancers first designed the robotic arm's movement sequences, then improvised with it. Dancers reflected on their transition from responding to designed movement sequences initially to improvisational expressions later on.

Participants generally did not perceive their interaction with the robotic arm as traditional communication, primarily because they were responsible for designing the robot's movements. This established a predictable structure that differed from human partnerships' dynamics and communication. As P1 described, "I reacted to it by moving my own body to it... I was just moving my positions in the whole area like on the dance floor to interact with it." This quote highlights how the improvisation was about spatial positioning rather than an ongoing dialogue, illustrating communication is design-based, not interactive.
However, the arm's predictability gave dancers a base for structured and spontaneous interactions. P5 noted, "So, every time you repeat it, it’s going to be the same. So, it gives me a very solid structure and then after that, I can have more freedom to dance." 
This structured framework offered dancers a foundation on which to base their improvisational movements, creating a balance between predictability and creativity.

A key theme was the shift from structured responses to improvisational movements. Initially, dancers engaged with the robotic arm's pre-designed motions, using them as a foundation for creativity. P4 noted, "When I dance with the robotic arms, the intention will be how I respond to my previous choreographed design. So, it is not purely improvisation." This reflects the premeditated nature of early interactions. However, as the process unfolded, dancers increasingly embraced improvisation. P1 described this progression: "At first, I wasn’t improvising much because the movements were my design. But as it progressed, I did more improvisational things." The robotic arm's predictable movements provided a structured framework that supported spontaneous exploration. Participants adapted to the arm's cues, integrating spatial awareness and dynamic positioning into their responses. P1 highlighted this integration: "I considered vertical and horizontal positions, repositioning myself to interact with the space dynamically." This interplay between control, reaction, and creativity enriched the dance experience, fostering a balance between structure and expression.

\subsubsection{Freedom of Movements in Individual Improvisation vs. Spatial Constraints in Group Improvisation.}

\begin{figure}[htbp]
  \centering
  \includegraphics[width=0.6\linewidth]{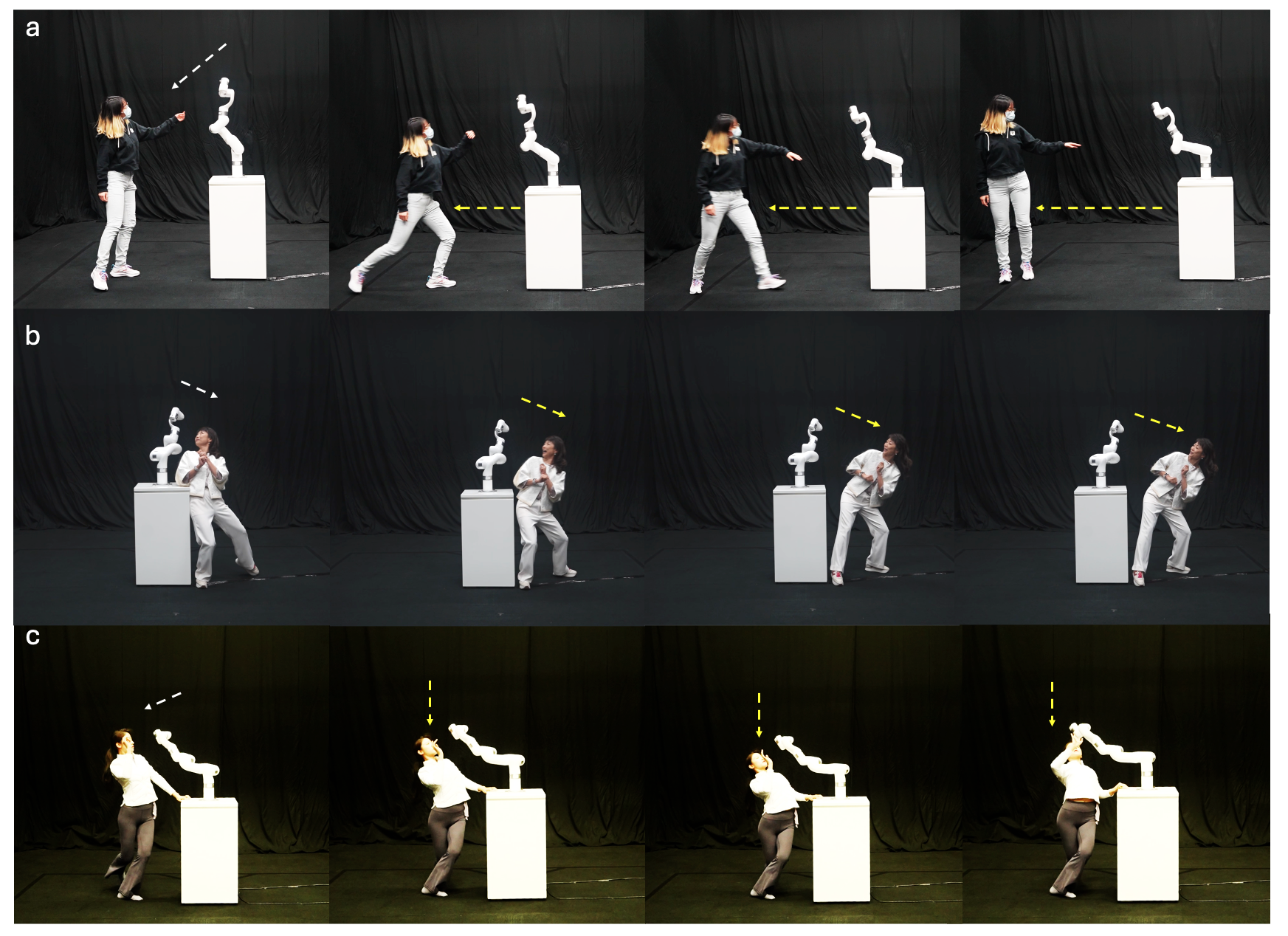}
  \caption{Different dance movement strategies for navigating potential hazards introduced by the robot arm. (a) P3 gets away from the robotic arm to avoid collisions. (b) P4 moves her upper body to avoid being hit when the robotic arm gets lower. (c) P9 lowers her body to avoid hazards in the robotic arm.}
  \label{fig:Fig9}
  \Description{Caption}
\end{figure}
Workshop 1 and Workshop 2 revealed notable differences between individual and group improvisational dance, especially in spatial dynamics and the contrasting predictability of robotic movements versus the adaptability of human dancers.

During individual improvisation, dancers found the robotic arm's predictable movements provided a stable framework for creative exploration. P5 remarked, "Every time you repeat, it’s going to be the same... It gives me a solid foundation and freedom to improvise." This predictability enabled dancers to experiment with spatial configurations and choreographic possibilities without unexpected disruptions.

In group improvisation (3-1), the presence of human partners introduced spatial constraints and the need for heightened awareness of others’ movements. P4 noted, "For humans, even if choreographed, reactions will vary because precision is less computationally determined." The unpredictability of human dancers required spontaneous adaptation, making interactions more dynamic yet challenging. P4 further observed, "In 3-1, we explore more. The robot is restricted, but humans can go up, down, and stretch the space."

Communication also diverged. Interactions with the robot were pre-set and design-driven, as P3 explained: "It’s not communication. It’s just we told it what to do." Conversely, human partners necessitated verbal and non-verbal coordination, offering greater spontaneity and creative dialogue. Some participants (P2, P3 and P4) appreciated the freedom and control of solo dances, others preferred group settings for their richness in expression and interaction. These differences highlight the distinct opportunities and challenges of human-robot collaboration in individual and group improvisation.

\subsubsection{Spatial Awareness and Navigation Behavior Around Hazards Involving Robot.}
Integrating robotic arms into dance required participants to balance creative expression with safety considerations. In the 1-1 setup, dancers showed heightened spatial awareness of the robotic arm's limitations and risks. P6 observed, "The robot doesn’t move in two directions... I have to create something that considers that because the robot can hit me, it will not stop."

In the 3-1 setup, saptial awareness extended to both the robot and fellow dancers. As P1 explained, "I was aware the robot wasn’t hitting me and where my other partners were." This awareness influenced choreographic choices, ensuring safe and fluid interactions (Fig.9). 
Across all workshops, dancers actively adjusted their positioning, showing that integrating non-humanoid robots into performance demands spatial cognition to avoid hazards while supporting expressive movement.

\subsubsection{Perspective Taking and Stage Awareness in Performance.}

Participants reflected on how robotic arms influenced their choreographic decisions in relation to stage presentation and audience perception. In 3-1 settings, dancers paid special attention to spatial balance, framing, and how the performance appeared from the audience's or camea's perspective. P1 explained, “I considered how the performance looks to the audience, especially in group dances where the entire stage and camera angles are important.” Dancers also explored the robotic arm’s aesthetic qualities, such as fluidity and precision, integrating these elements into their choreographic decisions and artistic vision.

\subsection{Perception and Presence of the Robotic Arm}
\begin{figure}[htbp]
  \centering
  \includegraphics[width=0.5\linewidth]{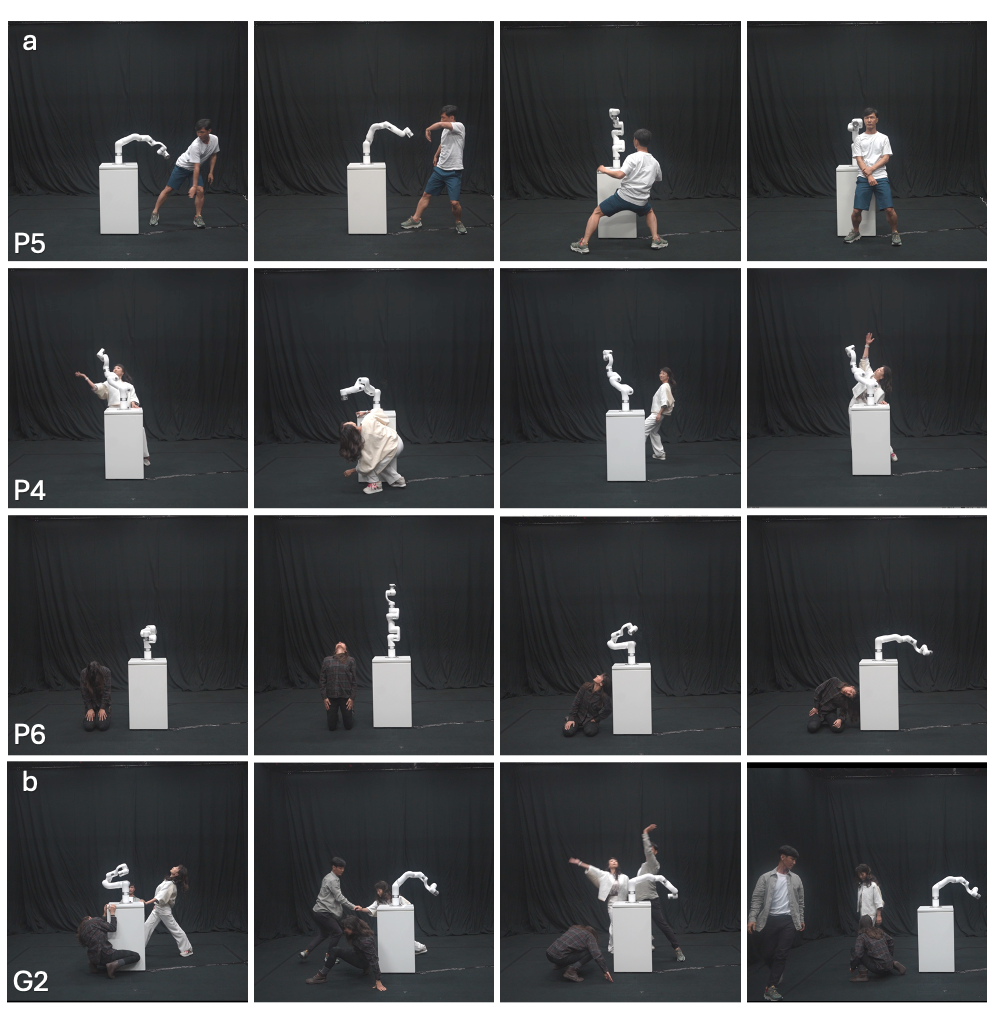}
  \caption{Screenshots of a. P4, P5 and P6 individual dance with the robotic arm. b. When doing group improvisational dance with the robotic arm, dancers put the robotic arm as the background and mostly interacted with the human dancers. }
  \label{fig:Fig10}
  \Description{Caption}
\end{figure}

\subsubsection{Perceiving Robot as the Background Element in Group Dance.}
Integrating a robotic arm in group dance settings significantly influenced dancers' perceptions and interactions. In 3-1 settings, dancers often shifted their focus towards human partners, relegating the robotic arm to a background role. P4 shared, “In 3-1, I completely ignore the robot, just interact with humans, and then the robot becomes my background.” This contrasts with 1-1 sessions, where the robot served as a central improvisational partner(Fig. 10).

Despite this, dancers sought to reintegrate the robotic arm into the choreography. As P4 explained, “When the interactions of humans came to an end, we tried to come back to the robot, try to respond, not left it behind.” This dynamic highlights the evolving interplay between human and machine, blending innovation with artistic expression.

\begin{figure}[htbp]
  \centering
  \includegraphics[width=0.6\linewidth]{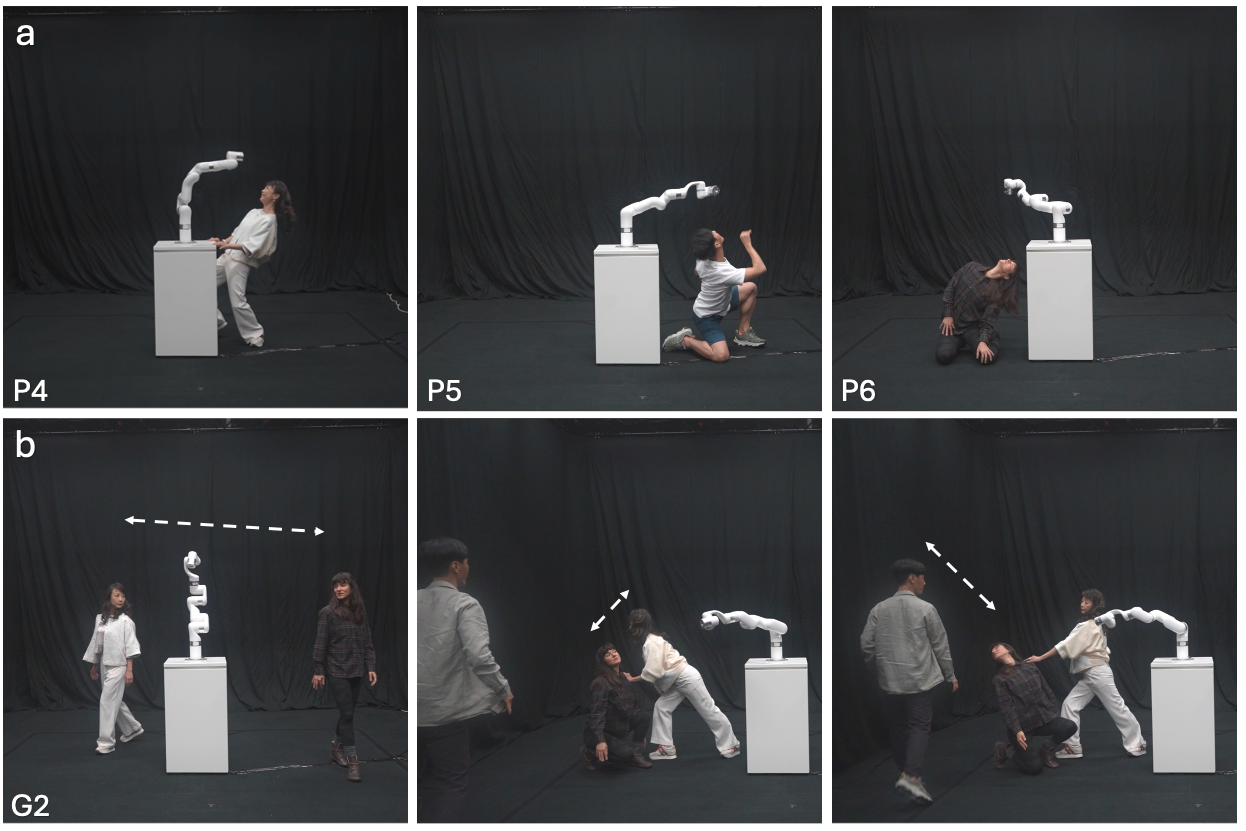}
  \caption{Screenshots showing attention allocation in individual vs. group improvisational dance. a. Dancers put their attention on the robotic arm mostly when individual dance. Figure a shows the examples from dancers P4, P5 and P6. In group improvisational dance, dancers put more attention on the human dancers than the robotic arm.}
  \label{fig:Fig11}
  \Description{Caption}
\end{figure}

\subsubsection{Attention Allocation in Individual vs. Group Improvisation: More Elements to Divide Dancer Attention.}
Moving from the individual (1-1) to group (3-1) improvisational sessions with the robotic arm shifts distinctly in where dancers direct their attention, alongside perceptions of the robotic arm and creativity. During individual sessions, dancers focused solely on the robot, treating it as a central and responsive partner. 

In contrast, the group setting introduced more cognitive complexity. P6 stated, "More elements to divide the attention. We can have an influence on others and at the same time be influenced by others. The robot has a fixed choreography, but human movements are not fixed; the way we looked each other, and we touch, we can influence and be influenced. The element of not knowing but also being able to shift the way of awareness and energy. The way the Energy is projected towards others can have influence, but project to the robot nothing will change." This dispersion of attention is exemplified by P5's reflection, emphasizing the need to allocate energy and relational dynamics between human partners and the robotic arm.

Additionally, dancers draw creative inspiration from the gestural vocabulary of the robotic arm, incorporating its articulations and postures into their choreographic dance. This artistic process is informed by their cognitive interpretation of the robotic arm's architecture and operation, with the trunk area serving as a hub of kinetic innovation and choreographic materials(P4, P6). This view finds resonance in their focused attention on the trunk or core area of the robotic arm, seen as a hub of kinetic vitality and choreographic integration. P4 elaborates on this idea, detailing their attentive focus on the middle and lower segments of the robotic arm, symbolizing their perception of it as an anchor for movement generation. Similarly, P6 metaphorically draws parallels between the robotic arm and her head, highlighting its "eye" as the source of motion. Despite technical constraints and safety considerations, dancers adeptly utilize the expressive potential of the robotic arm's gestures, using them as cues for choreographic innovation (P6).  During group setting, the perceptual significance of the robotic arm's trunk or core area persists as a main point for movement inspiration.

Comparing individual (1-1) and group (3-1) improvisations, dancers navigate an interplay of attention dynamics and creative ingenuity. Individual sessions allow for undivided focus on the robotic arm and its gestures. Group sessions, while introducing additional elements that claim a reallocation of cognition. Nonetheless, the perceptual centrality of the robotic arm's trunk or core area persists as a cornerstone of dancers' interaction and artistic expression across both contexts, revealing its pivotal role in choreographic exploration and improvisational interaction. This dynamic illustrates how improvisation in group contexts facilitates a balance between individual expression and collaboration. Participants emphasized that solo interactions with the robot deepened their personal connection and self-expression, group settings prompted shared creative decisions and a heightened sense of group identity through collaborative choreography.

\begin{figure}[htbp]
  \centering
  \includegraphics[width=0.7\linewidth]{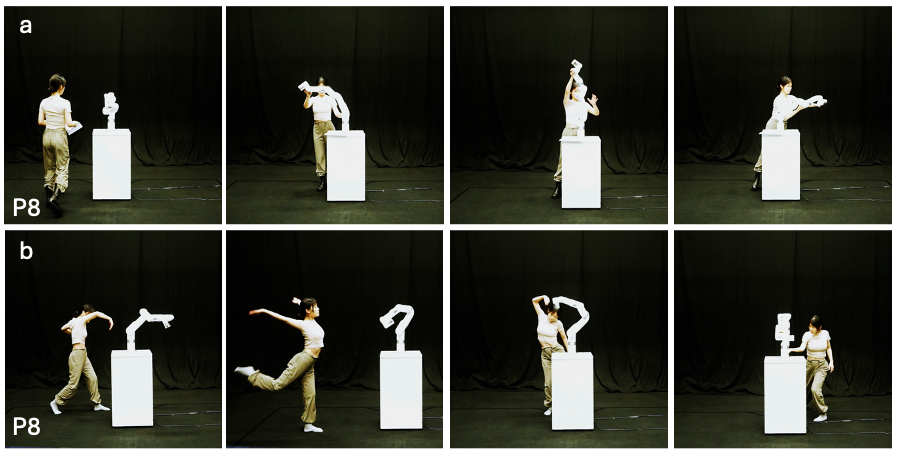}
  \caption{Relationship between dancer and the robotic arm changed from movement design to improvisational dance. Dancers perceived the robotic arm as a tool when doing movement design, while perceived the robotic arm as a partner when doing improvisational dancing. a. P8 did movement design for the robotic arm by manually moving the joints of the robotic arm and perceived it as a tool. b. P8 did individual improvisational dance with the robotic arm and saw it as a dance partner.}
  \label{fig:Fig12}
  \Description{Caption}
\end{figure}

\subsubsection{Context-specific Perception of Robot as Tool vs. Partner.}
Dancers’ perceptions of the robotic arm shifted significantly across different interaction contexts—from tool, to expressive medium, to dance partner. During the choreographic design phase in Workshop 1, most dancers saw the robotic arm as an extension of their intent. P4 stated, "Not much feeling. Because the robotic arm is manipulated by me. So, it's still an object in a way." The robotic arm was largely viewed as an extension of the dancers' creative intent, manipulated to explore choreographic possibilities.
Dancers projected contextual roles onto the robotic arm, as noted by P4, "Even though I tried to put in a character so if I want to be playful, then I become playful. If I want to be sad or emotional, then I become emotional. So that's how humans put in the emotional context on the robotic arm." Despite being perceived as a tool, the robotic arm became a medium for expression. For example, P6 perceived it as her finance, and P5 perceived the robotic arm as his student. 

During improvisational dance sessions, the robotic arm was perceived more as a dance partner. P2 described, "When I danced with it, at some point I felt it was like a real dancer, but the robot should not be so sudden, unlike a human." This indicates a shift in perception, with the robotic arm being seen as capable of mimicking a dance partner's presence. However, the perception varied among dancers. P4 mentioned, "When I am dancing, I do perceive it as a partner. Otherwise, I perceived it as a prop." The transition from tool to partner depended on the interaction context. P5's reflections highlighted this shifting role: "When I designed the movements, I used it as a prop, as a toy. But when dancing, it felt more like a partner...."

In group settings (Workshop 2), the robotic arm was predominantly perceived as a tool. Social dynamics made human partners the primary focus. P4 stated, "More humans to care about. It suddenly changed the relationships with the robot... 3-1, I completely ignore the robot just interact with humans and then the robot becomes my background." P5 also noted, "It is more like a set. As a prop to us. And we need to be part of it." Here, the robotic arm was perceived more as part of the stage than an active partner.

Comparisons with human dancers highlighted the robotic arm's limits in responsiveness. P1 stated, "Dancing with a human dancer, you will learn the movement from him/her. I did not have to tell what movement of him/her, just dance." This spontaneity and mutual learning were not present with the robotic arm, which was more predictable and controlled. In addition, P2 found dancing with robots more stimulating due to the heightened alertness required: "I think compared to dancing with humans, dancing with the robot is more exciting, because it's a kind of except for, they may hurt you. So this kind of makes your nerves try to keep an alarm. So I can feel even like after 20 seconds, my heartbeat is still faster." And P4 echoed this, "The difference will be because this robotic arm movement is designed by me. I know exactly how I design it. So, when I dance with the robotic arms, I am responding to my previous designs. Now, if I danced with a live person, sometimes very subtly, hit the movement change. I will react in a slightly different way so it's less predictable. So the robotic arm is much more predictable because they're completely designed by me so I know exactly when the movement will be." Moreover, P6 added, "Well, if I'm dancing with a person that is a professional dancer or something I will know this person is trained to be more aware of another person and not to injure. Because we're not fighting, we are dancing or something. But it depends, if I'm dancing with a child or with a person has a different notion of space and time or ability. So, this person also I need to also have a different awareness myself. So, the robot I mean, initially, it is it is something that will make me aware that it will not be aware of me, and it will not do anything to not hurt me. But if I stayed long enough, and I knew exactly what it was going to do, I could then play with that."

The perception of the robotic arm evolved throughout the workshops, from being viewed as a tool during movement design to being seen as a dance partner during improvisational sessions, and back to a tool in group settings. This evolution reflects the dynamic nature of collaboration dance with robot, highlighting both the potential and limitations of robotic arm in performance practices.

\begin{figure}[htbp]
  \centering
  \includegraphics[width=0.5\linewidth]{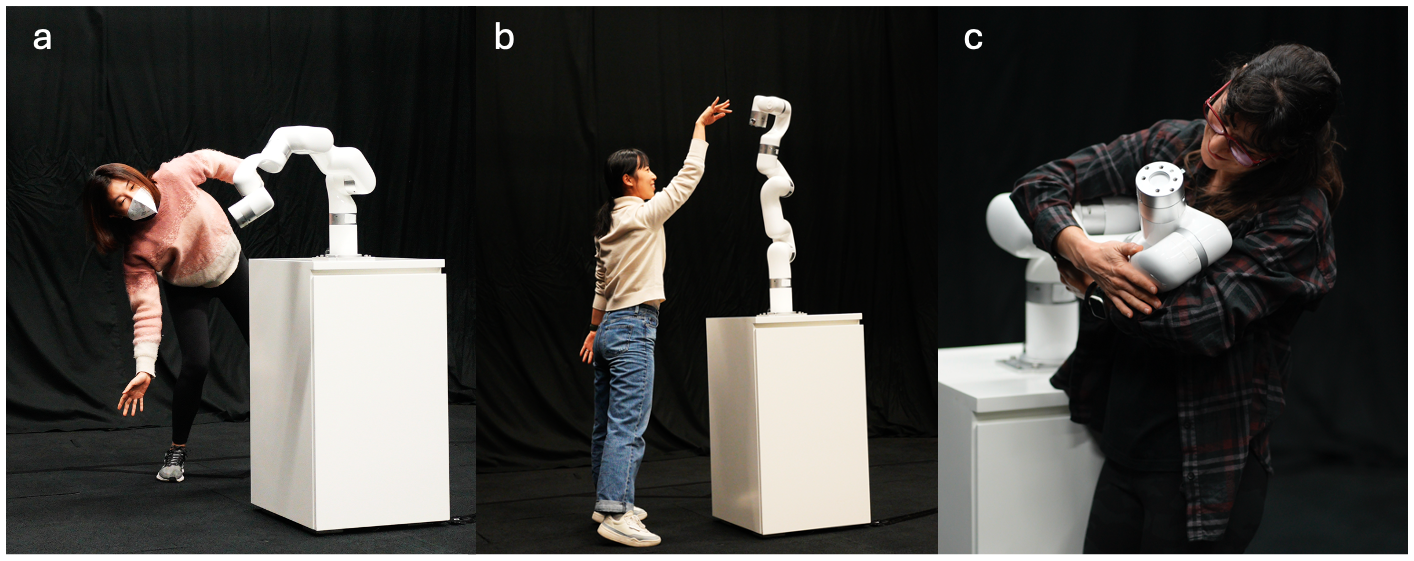}
  \caption{Examples of that dancers desire for more human-like interaction with the robotic arm. a. P1: Eye contact. b. P2: Touch. c. P6: Hug.}
  \label{fig:Fig13}
  \Description{Caption}
\end{figure}

\subsubsection{Dancer Desire for More Human-like Interactions with The Robotic Arm.}
Several dancers expressed a desire for the robotic arm to exhibit more human-like qualities-particularly responsiveness, mobility, and emotional presence-to better mirror the dynamics of human collaboration. P1 remarked, “I was expecting the robot to be more interactive...not stationary, but with four limbs, a body, and a head.” Similarly, P3 reflected: “I expected [it to] behave like the arm of a human.” These expectations highlight dancers’ longing for greater mobility, tactile interaction, and a humanoid form. (Fig.13)

Even with high precision and repeatablity, the robotic arm's limited fluidity and expressiveenss were noted as constraints. P2 compared the robot to a “perfect student,” appreciating its precision but noting its limitations compared to human partners: “When I dance with a human, there’s connection...the robot does not have that.” Dancers also observed the robot’s lack of fluidity and responsiveness. P3 noted its “sense of pause,” and P5 described its limited range of movement: “With humans, I can do many more possibilities...this one is with limitations.”

Participants often personified the robot as a snake, child, or dancer, reflecting their tendency to attribute human characteristics to it. This personification influenced their creative processes, highlights a need for building compelling narratives and thematic depth in their performances.

In general, dancers' insights show opportunities for future development in robotic choreography tools. Enhaving real-time adaptability, mobility, and emotional responsiveness may make robotic arms more compelling as creative collaborators-not just programmable tools.

\subsubsection{Absence of Human-like Features and Physical Connection: Limitations in Feedback.}
Dancers consistently pointed to the absence of human-like features and physical feedback as a barrier to collaboration and inspiration.
P1 noted, "Cannot inspire each other to do the movements like a human dancer," while P2 described it as a dancer that "can follow the rhythm...but [is] sudden, unlike a human." Without mutual inspiration or adaptability, dancers felt they were directing rather than collaborating with the robot. P1 added, "I'm deciding what the (robot) partner is going to move."

The robot's mechanical limitations-restricted joint motion and lack of full-body articulation-further shaped interaction. P5 observed, "With humans, I can do much more...this one (the robot) is with limitations," and P6 emphasized the lack of resilience, noting, "Robot won’t stop or fall, unlike human dancers."

Additionally, the robot's lack of real-time feedback posed challenges. P3 remarked, "The flexibility and speed of the robot are limited," highlighting how programming-based control lacked the immediacy of human feedback.

Despite these limitations, some participants appreciated the exploratory nature of interacting with the robot. P4 stated, "We’re trying to understand each other, trying to explore the possibilities." While the absence of human-like features and feedback posed challenges, the process still fostered creative exploration, hinting at potential for future improvements.

\subsection{Exploring Choreographer-Robotic Arm Collaboration}

\begin{figure}[htbp]
  \centering
  \includegraphics[width=0.7\linewidth]{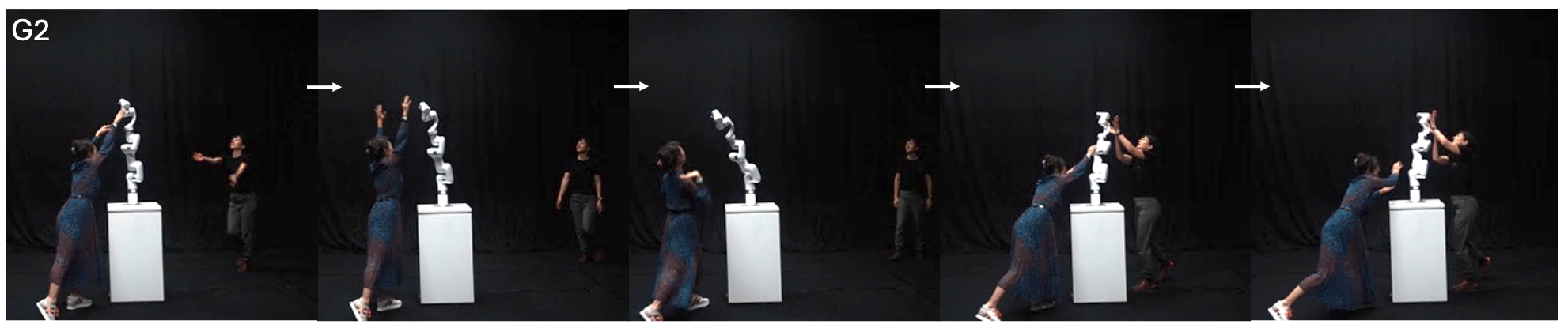}
  \caption{Examples of the robotic arm as a connection point between two dancers when choreography. In group 2 two dancers were choreographed to use the robotic arm as the medium for dancing.}
  \label{fig:Fig14}
  \Description{Caption}
\end{figure}

\subsubsection{The Robotic Arm as a Connection Point Between Two Dancers.}
In group choreography, the robotic arm served as a shared focal point, facilitating coordination, spatial structure, and visual balance. One choreographer emphasized its connective function: “The value of the robotic arm lies in giving the two dancers something in common” (G3).

In addition, choreographers perceive connections and interactions between dancers and robotic arm, regardless of whether the performance involves one or multiple dancers. As one expressed, "So yeah, so either one or two people come in, there are a lot of connections, interactions and relationships there" (G2). This common understanding of robotic arm choreography reflects choreographers leveraging technology to facilitate interplay between artistic expression and technological mediation within dance practices (Fig.14).

By strategically incorporating the robotic arm, choreographers enhanced spatial unity and balanced attention between human performers and the robotic elements, enriching the expressive and aesthetic qualities of the performance. 

\subsubsection{Choreographing for Human vs Robotic Dancers}

Choreographers identified key differences between creating movement for humans and the robotic arm. Human choreography typically focuses on narrative themes and relationships, whereas robotic choreography prioritizes form and imitation. One choreographer stated, "The difference is that I feel that in the past, choreography for people always had a theme and a fixed form" (G3), highlighting how robotic integration requires a shift in creative focus.

The theme setting creates a narrative or emotional context, guiding expressive exploration. Conversely, form setting emphasizes technical and spatial organization. As noted, "In the previous workshop 2, it was theme setting, but now we are setting the form" (G3). This shift determines the creative direction, focusing either on narrative depth or technical precision.

Choreographers face challenges in communicating with robotic arms, requiring physical force for manipulation. One choreographer explained, "When communicating with the robot, [you] need some force to achieve the movements" (G2). This claims to recalibrate traditional choreographic communication.

The transition from theme to form setting marked a significant shift. "In Workshop 2, it was theme setting. Now we are setting the form" (G3). This affected movement space and scheduling, making Workshop 3 more limiting. "The details are set, and it's more limiting" (G3). This highlights how choreographic approaches influence spatial dynamics and creative freedom.

Some choreographers used similar techniques for robotic arms as for human dancers. "I used similar choreography techniques for the robotic arm as I did for human dancers, emphasizing motif inspiration" (G1). This adaptability shows the challenges and possibilities of integrating robotic arms into dance. In summary, Workshop 3 illuminated the distinct and evolving considerations in choreographing for robotic arms versus human dancers. The transition from thematic to formal choreography marked a significant shift in creative approach and relational dynamics.

\subsubsection{Emotional Engagement and Spatial Arrangement Compared to Improvisation.}
\begin{figure}[htbp]
  \centering
  \includegraphics[width=0.7\linewidth]{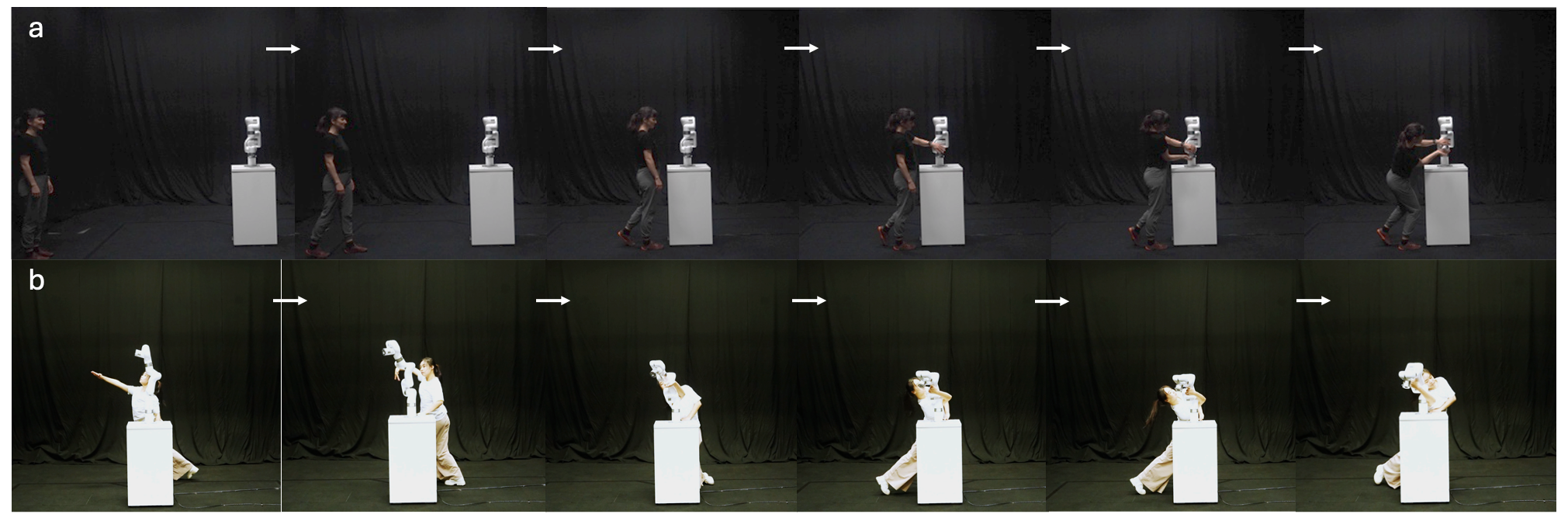}
  \caption{Different levels of emotion-based engagement during performance as directed by the choreographer. a. P6 was asked by the choreographer to approach the robotic arm slowly, and then touch it gently as her fiance. b. P7 was choreographed to hug the robotic arm as her dancer partner. }
  \label{fig:Fig15}
  \Description{Caption}
\end{figure}

In Workshop 3, dancers reported stronger emotional engagement during choreographed performances compared to the improvisations in Workshops 1 and 2. The structured choreography enabled dancers to fully express emotions, as G3 noted, "Choreography is more involved in emotion, and this action can better reflect the emotion." Additionally, choreographed movement patterns and spatial arrangements shaped interactions between dancers and the robotic arm, enhancing the flow and coherence of the performance. G3 remarked, "The influence is our movement space and scheduling space designed by the director."

Choreographers emphasized spatial arrangement and visual balance when integrating robotic arms, creating compelling compositions. Compared to improvisation, choreographing demanded more attention to spatial balance. One choreographer stated, "I will try to balance it as much as possible, according to the visual effect of balancing" (G3). This careful positioning enhanced aesthetic coherence and emotional impact. Another participant added, "Consider a transformation of its position to increase the overall performance effect, balance, and aesthetics" (G3).

These findings underscore the critical role of emotional expression and spatial arrangement in choreographed dances with robotic arms. Structured choreography not only deepened expressive potential but also enhanced the integration of robotic elements into compelling stage compositions.

%% file: sections/05-Discussion.tex
\subsection{Integrating Robotic Arm into Embodied Practice and Somatic Awareness}
Integrating robotic arms into dance practice fosters a unique interplay between technology and embodied performance, offering new avenues for choreographic exploration and somatic awareness. Introducing robotic arms into dance challenges the traditional boundaries between humans and machines, creating a hybrid form that necessitates new choreographic approaches. This integration is not merely about adding a technological component but involves rethinking the principles of dance itself. Previous work highlights the importance of designing interactive systems that are sensitive to the nuances of human movement, fostering a seamless blend between human dancers and robotic elements\cite{riviere2019capturing}\cite{riviere2021exploring}\cite{riviere2018dancers}\cite{alexander2023using}. Similarly, the study by Carlson et al.\cite{carlson2021mocap} stresses the significance of embodied interaction in developing intuitive interfaces that enhance the dancer's experience. Our findings resonate with these perspectives, demonstrating that effective integration requires choreographers to engage deeply with both the technological capabilities and the somatic experiences of dancers. In contrast to Riviere et al.\cite{riviere2019capturing}\cite{riviere2021exploring}, which focus on movement decomposition and interactive tools such as Knotation, our study distinctly investigates how non-humanoid robotic arms influence dancers' navigation of their somatic awareness. Unlike their emphasis on analyzing human movements for creative collaboration, we examine the adaptation process necessitated by the robot's mechanical and non-humanoid movements, which transform spatial and somatic practices. Furthermore, our findings show how workflows with non-humanoid robotic arms can inform creative tool design across domains. The iterative programming and physical manipulation motivate users to collaborate with interactive systems rather than just operate them \cite{dong2024dances}. Workshops' structured interaction between human dancers and robotic arms shows how such workflows can promote collaborative problem-solving and embodied learning in education, therapy, and other creative fields. Iterative design processes in choreography could be used to teach spatial reasoning or teamwork in team-based training. The findings suggest that robotic arms can spark new ways of working and innovation across diverse professional practices by emphasizing adaptive co-creation. Additionally, the spatial exploration and adaptability strategies identified in solo versus group settings could inform animation workflows in game design\cite{trajkova2024exploring}\cite{bautista2024understanding}, where teams collaboratively iterate on character movements in virtual environments. Similarly, iterative programming through demonstration, as employed in our movement design phases, could be adapted to collaborative prototyping in product design, enabling end-user creativity to shape technological behaviors. These insights demonstrate the potential for robotic systems to support not only artistic endeavors but also interdisciplinary applications that benefit from dynamic human-technology collaboration.

While earlier research has emphasized the role of inanimate objects as material extensions in dance\cite{bennett2020dancing} and how their appearance affects our perception\cite{friedman_designing_2021, friedman_what_2021, friedman_clothing_2024}, our study extends this discourse to non-humanoid robotic arms as dynamic collaborators. Unlike static objects, robotic arms bring mechanical precision and programmable adaptability to the performance space, creating new possibilities for choreographic interaction. This highlights the unique somatic and cognitive adjustments required when dancers engage with programmable, inanimate partners.

The shift from human-centric to robotic choreography brings a focus on form and precision, diverging from the thematic and relational nature of traditional dance. Our workshops revealed that choreographers had to navigate these formal constraints, adapting their methods to accommodate the robotic arm's limitations. These findings emphasize that solo improvisation with the robotic arm fosters deep personal engagement and self-expression, as dancers explore their own creative boundaries in response to the robot's movements. In contrast, group interactions balance individual contributions with collective expression, highlighting how group choreography enables dancers to co-create a shared narrative while maintaining their personal artistic identity. This contrasts with previous emphasis on flexibility and negotiation in technology design for dance, highlighting the need for adaptable choreographic strategies that can fluidly integrate technological elements without compromising artistic intent\cite{sullivan2023embracing}. Carlson et al. also emphasize the iterative design process in creating interactive systems that support the dancer's creative expression\cite{carlson2021mocap}. These approaches show the necessity for choreographers to develop hybrid methodologies that balance the technical demands of robotic choreography with the expressive richness of human dance. Furthermore, our research underscores the variation in dancers' somatic awareness as they transition between improvisational and choreographed environments while engaging with non-humanoid robots. This complements Mackay et al.\cite{riviere2018dancers}, which emphasizes the utilization of tools such as Knotation for depicting choreographic structures, by demonstrating the evolution of somatic awareness during real-time, embodied interactions with robotic arms.

In addition, familiarity with robotic movements emerged as a key factor in enhancing dancers' somatic awareness and interaction quality. In our study, dancers who were more familiar with the robotic arm's sequences could focus on personal expression and interaction, reducing cognitive load and facilitating a more intuitive performance. This aligns with Alaoui's findings that decomposing movement sequences into smaller, manageable parts can help dancers internalize complex phrases, thereby enhancing somatic engagement\cite{riviere2019capturing}\cite{riviere2021exploring}. Carlson et al.'s work further supports this, suggesting that embodied interaction with technology can deepen the dancer's kinesthetic awareness, leading to more fluid and responsive performances. These insights highlight the importance of incorporating pedagogical strategies that emphasize familiarity and repetition in robotic choreography.

Our study suggests that choreographed dances with robotic arms can elicit significant emotional involvement from dancers, provided the choreography balances structure and expressive freedom. Prior research demonstrates that structured choreography can serve as a stable foundation for emotional exploration, as seen in her studies on movement decomposition and its impact on learning and performance quality\cite{riviere2021exploring}\cite{riviere2018dancers}. Another work also notes that narrative and emotional depth in interactive performances are enhanced when the technology supports rather than dictates the artistic vision\cite{dipaola2016movement}. By designing robotic choreography that allows for both precision and emotional expressiveness, choreographers can create compelling narratives that resonate deeply with both dancers and audiences.

Effective integration of robotic arms into dance requires attention to spatial arrangement and visual balance. Our workshops highlighted the importance of strategic positioning to create visually compelling compositions. Alaoui's research on technology design in dance emphasizes the necessity of considering spatial dynamics to enhance the overall aesthetic and narrative coherence of performance and emphasize the role of spatial awareness in interactive systems, advocating for designs that facilitate a harmonious blend of human and robotic elements \cite{sullivan2023embracing}. Similarly, This involves a delicate interplay between dancers and robotic components, ensuring that the technological presence enhances rather than detracts from the performance's visual impact.

\subsection{Rethinking the Relationship and Perception between Dancer and Robotic Arm}

\subsubsection{Comparison: Human and Robotic Arm Dancers.}
The robotic arm presents unique challenges and opportunities compared to human dance partners. Participants noted the absence of verbal communication and the need for alternative feedback forms, such as visual and haptic cues. It has been emphasised that the importance of non-verbal communication in choreographic movement\cite{singh_choreographers_2011}. The robotic arm’s precision and consistency were valued, yet its inability to improvise and respond organically was a limitation. Future research should focus on enhancing the robot's real-time responsiveness to mimic the adaptability of human partners. Compared to Alaoui et al.\cite{alaoui2021rco}, which examines creativity facilitated by digital tools, our findings concentrate on real-time, embodied interactions with non-humanoid robots. While their research focuses on visual representation, ours examines physical and somatic adaptations to the robot's mechanical limitations, enhancing understanding of co-creative processes in performance.

\subsubsection{Emotional Communication in Robotic Arms.}
The robotic arm's limited emotional expression capacity is a critical improvement area. Participants found the robot's movements mechanical and lacking the subtlety of human emotion. The importance of emotional design in creating meaningful interactions with technology has been highlighted\cite{norman2004}. Enhancing the robot's ability to convey a broader range of emotions through sophisticated movement algorithms\cite{li_nice_2023,xu_cohesiveness_2022} would improve its role in dance, allowing it to participate more fully in the performance's emotional, playful, and narrative aspects\cite{lc_machine_2020}. Riviere et al.\cite{riviere2021exploring} emphasize the narrative capabilities of digital tools in the creation of dance. Our research expands upon this by examining how the absence of emotional feedback from non-humanoid robots alters choreographers' creative narratives, compelling them to emphasize structure and deliberate designs rather than spontaneous emotional connections.

\subsubsection{Robotic Arm: Partner or Tool? Lead or Follow? Exploring its Role in Dance Performance.}
The dual role of the robotic arm as both a partner and a tool was a recurring theme. Some dancers viewed the robot as an extension of their body, while others saw it as a separate entity. It has been discussed that duality, where technology can function as both an instrument and a collaborator\cite{suchman2007}. This dual perception influences how dancers interact with the robotic arm and integrate it into their performances. Prior study focused on the exploration of motion and object interaction\cite{sheets1979movement}, argue that inanimate elements in performance serve as either extensions of the body or as external stimuli for creative action. Similarly, our findings reveal that the robotic arm functions as both a creative partner and a functional tool. Unlike static objects, its programmable movements provide dancers with a dynamic framework, bridging the gap between interactive materiality and performative agency. Our findings highlight how dancers adapt to the robotic arm as a dual agent compared with prior work\cite{riviere2018dancers}. This involves not only its tool-like predictability but also its integration as a co-dancer with its own mechanical idiosyncrasies.

The dynamic of leading and following in dance with a robotic partner presents unique challenges. Participants desired more fluid and responsive interactions where the robot could adapt to their movements in real-time. Virtual reality applications that adapt seamlessly to user inputs could inform the development of more responsive robotic dance systems\cite{hong2024dance}. Developing a robotic arm that can respond dynamically and adapt to the dancer's movements could enhance the collaborative aspect of the dance, allowing for more spontaneous and organic performances.

The integration of robotic arms into dance opens new avenues for narrative and storytelling. Participants highlighted the potential for robots to embody characters or abstract concepts, adding complexity to the performance. The potential of integrating virtual elements into physical performances has been stressed to enhance storytelling\cite{billinghurst2001}. Incorporating robots into dance allows artists to explore new narrative structures and create performances that blend human and robotic elements innovatively.

\subsection{Dancers Adapting Their Movements to The Robotic Arm}
The integration of robotic arms into dance practices requires dancers to adapt to movements that significantly differ from theirs. This adaptation is a complex process that involves both cognitive and physical adjustments, as well as a deep engagement with the technological elements. Our findings provide a detailed account of this adaptation process. One critical aspect of adapting to robotic movements is the necessity for dancers to reinterpret and recontextualize their understanding of movement. Previous work on capturing movement decomposition supports this idea by emphasizing the importance of breaking down complex movements into manageable components for teaching and learning in contemporary dance\cite{riviere2019capturing}. This process is equally applicable when dancers interact with robotic arms, as it enables them to understand, prototype\cite{lc_sit_2024, lc_now_2021}, and internalize the mechanical and pre-programmed nature of robotic movements. Our findings align with the emphasis on decomposition\cite{riviere2019capturing}, as choreographers in our workshops highlighted the need to dissect robotic movements to better integrate them into human choreography.

Moreover, the study on the role of artifacts in collective dance sheds light on how external elements, such as robotic arms, can influence dance practices\cite{riviere2021exploring}. Carlson et al. argue that artifacts serve as crucial points of reference and interaction, shaping the spatial and temporal dynamics of performance. This perspective is particularly relevant to our findings, where robotic arms acted as both partners and props, necessitating that dancers continually adapt their movements to the robotic elements. The dynamic interplay between human and machine in our workshops echoes Carlson et al.'s findings, highlighting the importance of artifacts in contemporary dance.

Furthermore, how dancers learn to dance has been explored to emphasize kinaesthetic learning and the importance of iterative practice\cite{riviere2018dancers}. Our workshops revealed that dancers adapted to robotic movements through repeated interaction and feedback, echoing Rivière et al.'s emphasis on the iterative nature of learning and adaptation in dance. 

In summary, adapting to movements that differ from a dancer's own involves incorporating traditional choreographic principles with innovative strategies tailored to robotic elements. Insights from previous works show the importance of movement decomposition\cite{riviere2019capturing}, kinaesthetic learning\cite{riviere2018dancers}, and the role of artifacts\cite{riviere2021exploring} in facilitating collaboration performance with technology. These concepts are crucial for choreographers and dancers to navigate the challenges of integrating robotic arms into dance, enriching the creative potential and performative experience of dance.

\subsection{Limitations}

Integrating robotic arms into dance practice presented several challenges and limitations. The spatial constraints imposed by the robotic arm's setup significantly influenced the dancers' movements. The fixed position and limited range of motion of the robotic arm forced dancers to adapt their improvisations to a confined space, which restricted their ability to perform more dynamic and expansive movements. This limitation required dancers to rethink their designs and adapt to the robotic arm’s mechanical constraints, often leading to simplified and less fluid performances.

The small sample size of nine participants limits the generalizability of our research's findings to the broader dance community. While our participants were professional dancers, they did not fully represent the diversity of dance forms, such as cultural dances (e.g., Flamenco or Bharatanatyam) or acrobatic aerial styles (e.g., pole dancing or aerial silks). A larger and more diverse sample could reveal how different movement vocabularies and traditions affect interactions with robotic systems. For instance, cultural dancers might adapt uniquely due to their storytelling or prop-based practices, while ballet dancers' structured techniques could influence their ability to work with non-humanoid designs. Expanding the participant pool in future research would provide deeper insights into how robotic systems foster creativity and collaboration across varied artistic and cultural contexts.

Although participants self-reported no prior experience with the robotic arm, some participants may have had experience dancing with other inanimate objects (e.g., chairs, umbrellas). While these experiences were not explicitly addressed in the interviews, such prior interactions with non-living objects could have influenced their approach and perceptions of the robotic arm. This limitation should be considered when interpreting the results, as prior experience with inanimate objects may have shaped their responses and interactions in ways not fully captured by the study.

Mechanical limitations such as joint resistance and movement speed posed additional challenges. Dancers found it difficult to synchronize their movements with the music, as the robotic arm could not keep pace with faster tempos or execute more complex movements smoothly. These mechanical constraints necessitated a careful balance between the robot's capabilities and the choreographic intentions, often requiring significant adjustments to the original dance routines.

The static nature of the robotic arm also posed a significant constraint. Unlike human dancers who can move fluidly across the stage, the robotic arm's fixed position limited the spatial dynamics of the performance. Dancers had to navigate around the stationary robot, which influenced their movement patterns and limited their ability to engage in more spontaneous and interactive choreographic elements.

Technological challenges further compounded these issues. The pre-programmed nature of the robotic arm restricted the dancers' ability to make spontaneous adjustments during the performance, limiting the improvisational aspect of the dance. Safety concerns were also prevalent, as dancers had to remain cautious to avoid collisions with the robot, which could not stop quickly if it moved unexpectedly. This preventive approach sometimes hindered the natural flow of the performance.

Additionally, the absence of human-like features and the lack of physical feedback from the robotic arm impacted the dancers' ability to form a meaningful connection with their mechanical partners. Unlike human dancers who can provide real-time feedback and adapt dynamically, the robotic arm's pre-set movements lacked responsiveness, which limited the depth of the interaction. This absence of mutual inspiration led to a more directive approach, where dancers felt they were instructing the robot rather than collaborating with it.

Despite these constraints, some dancers recognized the unique capabilities of the robotic arm, such as its precise rotational movements, which offered new possibilities for choreography. However, the overall experience highlighted the need for further advancements in robotic technology to enhance the fluidity, responsiveness, and interactive potential of robotic arms in dance performances. Future research should focus on addressing these limitations to facilitate a more seamless and integrated collaboration dance with non-humanoid robot.

\subsection{Exploring Applications in Kinaesthetic Development and Dance Education: Interdisciplinary Collaboration and Knowledge Transfer}
The integration of robotic arms into dance reveals new opportunities for kinaesthetic learning and collaborative movement design, advancing interdisciplinary collaboration between robotics, dance, and technology. Our study highlights how non-humanoid robotic arms encourage dancers to adapt to mechanical constraints, fostering innovative approaches to movement creation. This supports kinaesthetic education by enhancing dancers' ability to refine techniques and balance artistic intent with technical limitations.

\textbf{Adapting Movements to Robotic Constraints}
Dancers iteratively adjusted their movements to align with the robotic arm's capabilities, such as 360-degree rotation and static positioning. This process created a dynamic learning environment, enabling the exploration of expressiveness within defined mechanical boundaries. These adaptations illustrate how robotic tools can encourage both novice and expert dancers to engage in creative problem-solving.

\textbf{Enhancing Performance and Audience Engagement}
The robotic arm introduced unique opportunities for performance, offering consistent and precise movements that enhanced choreography's aesthetic and emotional impact. Performances emphasized audience engagement, evoking active participation and emotional connection through innovative interaction with the robotic arm.

\textbf{Future Directions and Collaboration}
The study underscores the potential for further development in robotic flexibility, emotional responsiveness, and expanded movement capabilities. These advancements could deepen the role of robotics in dance, enriching creative expression and collaborative practices. Moreover, interdisciplinary collaboration—uniting robotics, computer science, and performing arts—fosters innovation and knowledge transfer, enabling broader applications beyond dance.

%% file: sections/06-Conclusion.tex
This study examined how the non-humanoid robotic arm influences creative movement, choreography, and collaboration in dance. Across three workshops, dancers engaged with the robot in solo and group settings, revealing shifts in spatial awareness, interaction dynamics, and role perception. One-to-one settings fostered more fluid, intimate engagement, while group contexts introduced divided attention and reduced the robot’s prominence. This study shows how non-humanoid robots challenge choreographers to rethink group performance structures and inspire new collaborative approaches to movement design.


Our findings show the need for dancers and choreographers to adapt traditional methods to incorporate robotic arms effectively, blending dance techniques with new strategies tailored to the robot's capabilities and limitations. Familiarity with the robotic sequences enhanced dancers' somatic awareness and interaction quality, suggesting the importance of iterative practice and deep engagement with technological elements. While current limitations include the lack of real-time feedback and emotional responsiveness, our study points toward design opportunities for more adaptive, expressive robotic systems.  

Beyond dance, these insights inform the design of interactive robotic systems for movement-based learning and co-creative collaboration. The observed iterative adaptation parallels processes in education and design, where structured yet flexible human-technology interactions enhance creative exploration. Our research highlights how robotic arms enhance creative workflows in dance and offers a framework for integrating interactive systems into domains like education and collaborative design, where users co-create with technology. These insights demonstrate how technology can drive innovation in dance, emphasizing the potential for interdisciplinary collaboration to augment human creativity in novel and transformative ways.
